\documentclass{article}
\usepackage{microtype}
\usepackage{graphicx}
\usepackage{subfigure}
\usepackage{booktabs} 
\usepackage{hyperref}
\usepackage{subcaption}
\usepackage{caption}

\usepackage{wrapfig}
\usepackage[accepted]{icml2024}

\usepackage{amsmath}
\usepackage{amssymb}
\usepackage{mathtools}
\usepackage{amsthm}
\usepackage{multirow}
\usepackage{multicol}
\usepackage{float}

\theoremstyle{plain}

\theoremstyle{definition}

\theoremstyle{remark}

\usepackage[textsize=tiny]{todonotes}

\renewcommand{\thefigure}{Figure \arabic{figure}}
\renewcommand{\figurename}{}

\renewcommand{\thetable}{Table \arabic{table}}
\renewcommand{\tablename}{}

\icmltitlerunning{SemioLLM}
\begin{document}
\onecolumn
\icmltitle{SemioLLM: Evaluating Large Language Models for Diagnostic Reasoning from Unstructured Clinical Narratives in Epilepsy}


\begin{icmlauthorlist}
\icmlauthor{Meghal Dani}{uni,MLCluster,hertieAI}
\icmlauthor{Muthu Jeyanthi Prakash}{uni,uniclinic,hertieAI}
\icmlauthor{Filip Rosa}{uniclinic,hertiebrain}
\icmlauthor{Zeynep Akata}{helmholtz,tum}
\icmlauthor{Stefanie Liebe}{uniclinic,hertiebrain,hertieAI}
\end{icmlauthorlist}

\icmlaffiliation{uni}{University of Tübingen, Tübingen, Germany}
\icmlaffiliation{tum}{Technical University of Munich, Munich, Germany}
\icmlaffiliation{helmholtz}{Helmholtz Munich, Munich, Germany}
\icmlaffiliation{uniclinic}{Dept. of Neurology and Epileptology, University Clinic Tübingen,Germany}
\icmlaffiliation{hertiebrain}{Hertie Institute for Clinical Brain Research, Tübingen, Germany}
\icmlaffiliation{MLCluster}{Machine Learning in Science, Excellence Cluster Machine Learning, Tübingen University, Germany}
\icmlaffiliation{hertieAI}{Hertie Institute for AI in Brain Health (Hertie AI)}

\icmlcorrespondingauthor{Meghal Dani}{meghal.dani@uni-tuebingen.de}
\icmlcorrespondingauthor{Stefanie Liebe}{stefanie.liebe@uni-tuebingen.de}
\icmlkeywords{AI for Science, AI in healthcare, Machine Learning, LLMs, Epilepsy, Neuroscience}
\printAffiliationsAndNotice{} 
\vskip 0.3in
\begin{abstract} 
Large Language Models (LLMs) have been shown to encode clinical knowledge. Many evaluations, however, rely on structured question-answer benchmarks, overlooking critical challenges of interpreting and reasoning about unstructured clinical narratives in real-world settings. In this study we task eight Large Language models including two medical models (GPT-3.5, GPT-4, Mixtral-8x7B, Qwen-72B, LlaMa2, LlaMa3, OpenBioLLM, Med42) with a core diagnostic task in epilepsy: mapping seizure description phrases—after targeted filtering and standardization—to one of seven possible seizure onset zones using likelihood estimates. Most models yield results that often match the ground truth and even approach clinician-level performance after prompt engineering. Specifically, clinician-guided chain-of-thought reasoning leading to the most consistent improvements. Performance was further strongly modulated by clinical in-context impersonation, narrative length and language context (13.7\%, 32.7\% and 14.2\% performance variation, respectively). However, expert analysis of reasoning outputs revealed that correct prediction can be based on hallucinated knowledge and inaccurate source citation, underscoring the need to improve interpretability of LLMs in clinical use. Overall, \emph{SemioLLM} provides a scalable, domain-adaptable framework for evaluating LLMs in clinical disciplines where unstructured verbal descriptions encode diagnostic information. By identifying both the strengths and limitations of LLMs, our work contributes to testing the applicability of foundational AI systems for healthcare.
\end{abstract}

\section{Introduction}\label{sec:intro}
Large Language Models (LLMs) have shown significant potential in leveraging clinical knowledge on structured question answering (Q\&A) datasets such as MedQA~\cite{jin2021medqa}, PubMedQA~\cite{jin2019pubmedqa}, MedMCQA~\cite{pal2022medmcqa} and BioASQ-QA~\cite{krithara2023bioasq} in multiple medical domains~\cite{sarvari2024systematic, savage2024diagnostic, singhal2023large, singhal2023towards, van2024potential}. 
While Q\&A set-ups are advantageous as they provide a clear ground-truth for model testing, they oversimplify clinical decision-making ~\cite{hager2024evaluation}, which often relies on extracting crucial diagnostic information from unstructured patient interviews containing complex, irrelevant and everyday language~\cite{alaa2025medical, european2023esr, raji2025s}. While LLMs have shown remarkable capabilities in extracting meaningful information from unstructured text for a wide range of downstream tasks in other domains~\cite{bommasani2021opportunities}, their ability to do so in clinical contexts remains poorly understood. Thus, clinical narratives from patients provide crucial diagnostic information to be leveraged by LLMs especially when structured input is limited or absent \cite{van2024potential, yan2024large}. However, it is an open and largely unexplored question of how well LLMs can extract and interpret clinically meaningful information from unstructured clinical narratives to support real-world diagnostic reasoning.

Neurological disorders, such as epilepsy, provide a particularly compelling use case to explore this question, as behavioral and sensory symptoms can often be directly linked to underlying brain pathologies. In epilepsy, clinicians routinely base clinical decisions on patient and witness accounts of seizure manifestations—known as \emph{semiology}~\cite{nowacki2017evaluation, tufenkjian2012seizure}. Especially during early diagnostic evaluations, correctly interpreting seizure symptoms is crucial for guiding follow-up procedures such as brain imaging, EEG and surgical planning ~\cite{beniczky2022seizureilae, luders2006epileptogenic}. Seizure descriptions contain diagnostically relevant information about the seizure origin in the brain, and help classifying seizure types and syndromes. For example, repetitive chewing, swallowing, or lip-smacking strongly indicate temporal lobe involvement~\cite{wiebe2001randomizednejm}, while excessive limb movements or pelvic thrusting are indicative of frontal lobe seizures~\cite{luders2006epileptogenic}. Accurate localization of the seizure onset zone (SOZ) is particularly important for patients with drug-resistant epilepsy, where surgical resection of the SOZ remains the only potentially curative treatment option~\cite{sisodiya2007dre, wiebe2001randomizednejm}.
 
In this study, we develop a structured and automated evaluation framework, \emph{SemioLLM}, that benchmarks LLMs' ability to extract and translate diagnostically relevant information from seizure descriptions into probabilistic seizure locations in the brain. Using an annotated database linking over 1,200 seizure descriptions to seizure foci~\cite{semio2brain}, we evaluate eight LLMs, including proprietary and open-source models (GPT-3.5~\cite{brown2020language}, GPT-4~\cite{achiam2023gpt4}, Mixtral-8x7B~\cite{jiang2024mixtral}, Qwen-72B~\cite{bai2023qwen}, LlaMa2 70B~\cite{touvron2023llama2}, and LlaMa3 70B~\cite{dubey2024llama3}, OpenBioLLM, Med42~\cite{christophe2024med42}). We systematically analyze model accuracy, confidence, calibration and reasoning, benchmarking their outputs in comparison to evaluation by a clinical domain expert. 

Our study reveals several key insights: (i) We demonstrate that LLMs can predict seizure onset zones (SOZ) based on unstructured seizure descriptions, significantly outperforming chance-level predictions. Importantly, with refined prompting techniques, for example chain-of-thought (CoT) reasoning,  their accuracy substantially improves and approaches clinician-level performance. (ii) Many models exhibit reasonable trustworthiness, as assessed through an entropy-based confidence measure, where confidence similarly improves with prompt-engineering. Notably, GPT-4 and Mixtral-8x7B demonstrated an optimal balance of accuracy \textit{and} confidence. (iii) Through an extensive manual reasoning assessment conducted by a domain specific clinical expert, we identify that GPT-4 demonstrates superior capabilities in integrating domain knowledge, clinical inference, and evidence verification. Mixtral-8x7B, while competitive in text comprehension, exhibits notable limitations in reasoning and accurate source citation, underscoring areas for future model refinement. (iv) Our analysis identifies key factors that influence LLM diagnostic performance in processing seizure descriptions. First, we observe a U-shaped relationship between description length and accuracy, with both very short and highly detailed narratives yielding better results than those of moderate length. Second, prompting models to impersonate clinical experts markedly improves both accuracy and confidence, suggesting better alignment with domain-specific reasoning. Third, our multilingual evaluation shows that while English-trained models perform well even when processing non-English clinical narratives, accuracy declines for different prompt languages, underscoring the need for targeted multilingual user adaptation. 

In summary, our study provides a systematic and in-depth investigation of LLMs in epilepsy diagnostics based solely on verbal symptom descriptions, identifying prompt strategies and expert impersonation as the most significant factors that improved diagnostic accuracy on average by 10\% and 13.8\%, respectively. Unlike existing structured Q\&A evaluations, SemioLLM demonstrates how LLMs can translate unstructured clinical narratives into probabilistic diagnostic decisions, and identifies conditions under which both accuracy and confidence can be improved. Our framework provides a practical guideline for similar deployments and is easily transferable to other clinical specialties where symptom descriptions inform diagnostic decisions, potentially improving early diagnosis and treatment planning for patients with complex neurological and other medical conditions.

\begin{figure*}[]
\centerline{\includegraphics[width=\textwidth]{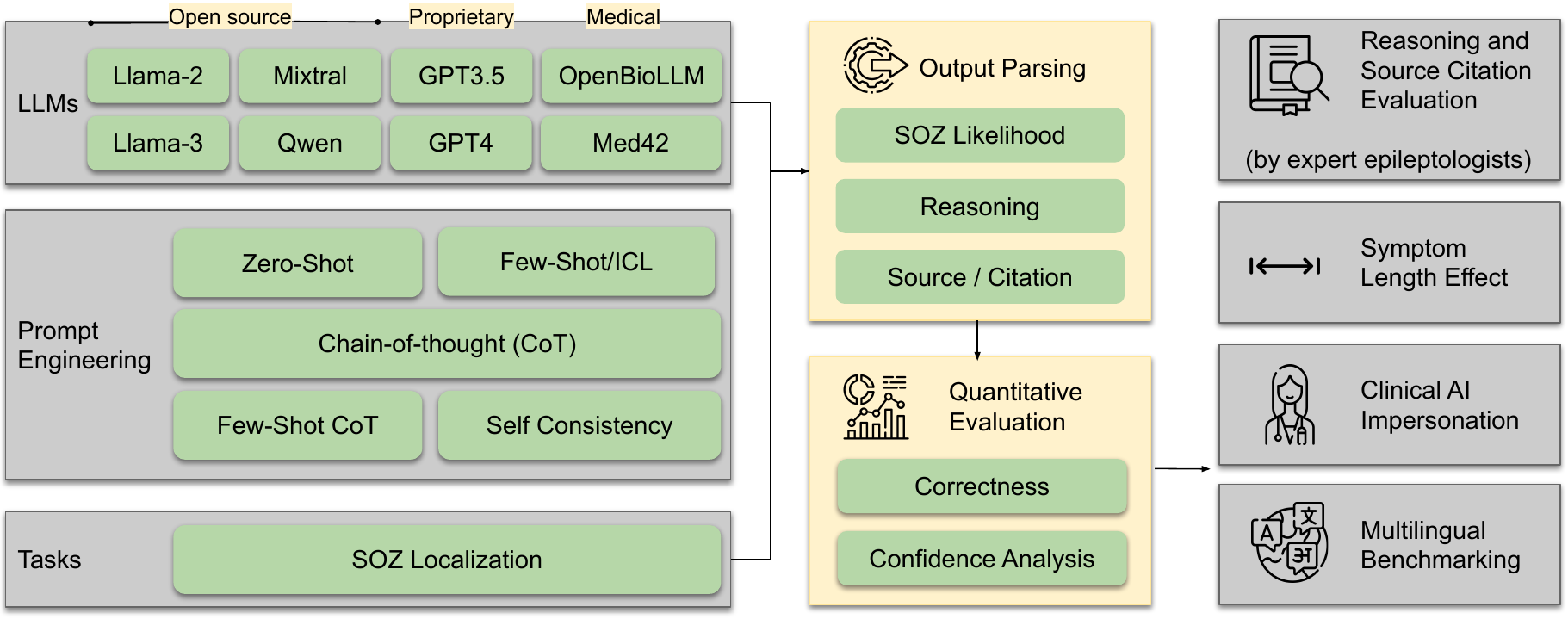}}
\caption{Overview of \emph{SemioLLM}: We consider eight open-source and proprietary LLMs and evaluate them across five standard prompt styles for the task of SOZ localization. Model outputs include likelihood estimates of seven major brain regions, reasoning and source citations and are evaluated for accuracy and confidence. The best performing models are examined in more detail with respect to a) task comprehension, logical reasoning, knowledge retrieval, clinical safety and source citation verification, (b) impact of symptom description length, c) in-context clinical impersonation, and d) multilingual alignment and understanding. The icons used in the creation of figure are sourced from \href{https://www.flaticon.com/}{Flaticon}. 
}
\label{fig:main}
\vspace{-3mm}
\end{figure*}

\begin{figure*}[!ht]
\centerline{\includegraphics[width=0.98\textwidth]{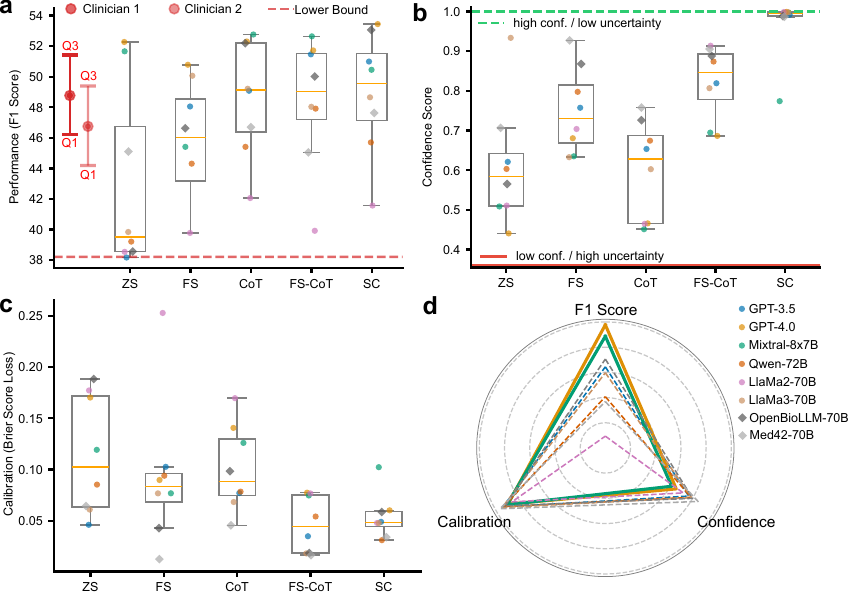}}
\caption{Comparison of LLMs (6 general, 2 medical) and impact of prompt engineering strategies [Zero-Shot (ZS), Few-Shot (FS), ZS-Chain-of-Thought (CoT), FS-CoT and Self Consistency (SC)] on performance, confidence and calibration. (a) Mean F1 scores from 999 bootstrap samples of n=1,269 symptoms (n=8 models per prompt style) F1 scores for all models obtained by bootstrapping. Advanced prompt styles achieve performance comparable to clinicians and at par with naive classifier (38.2\%,red dashed line) (b) Mean confidence scores (0=low, 1=high) from 999 bootstrap iterations of n=1,269 symptoms (n=8 models per prompt style). In-context learning improves confidence, with FS and FS-CoT showing largest gains (c) Calibration via Brier Score Loss from n=1,269 symptoms (n=8 models per prompt style); lower indicates better calibration. FS-CoT and SC show best calibration. (d) Multidimensional performance visualization comparing model correctness, confidence, and calibration metrics, with solid lines representing the best-performing models. Box plots in a-b show the interquartile range (IQR; 25th–75th percentile) with the median indicated by the orange line. Points represent individual models (circles: general LLMs; diamonds: medical LLMs)
} 
\label{fig:corr_result}
\vspace{-3mm}
\end{figure*}

\begin{figure*}[!ht]
\centerline{\includegraphics[width=0.98\textwidth]{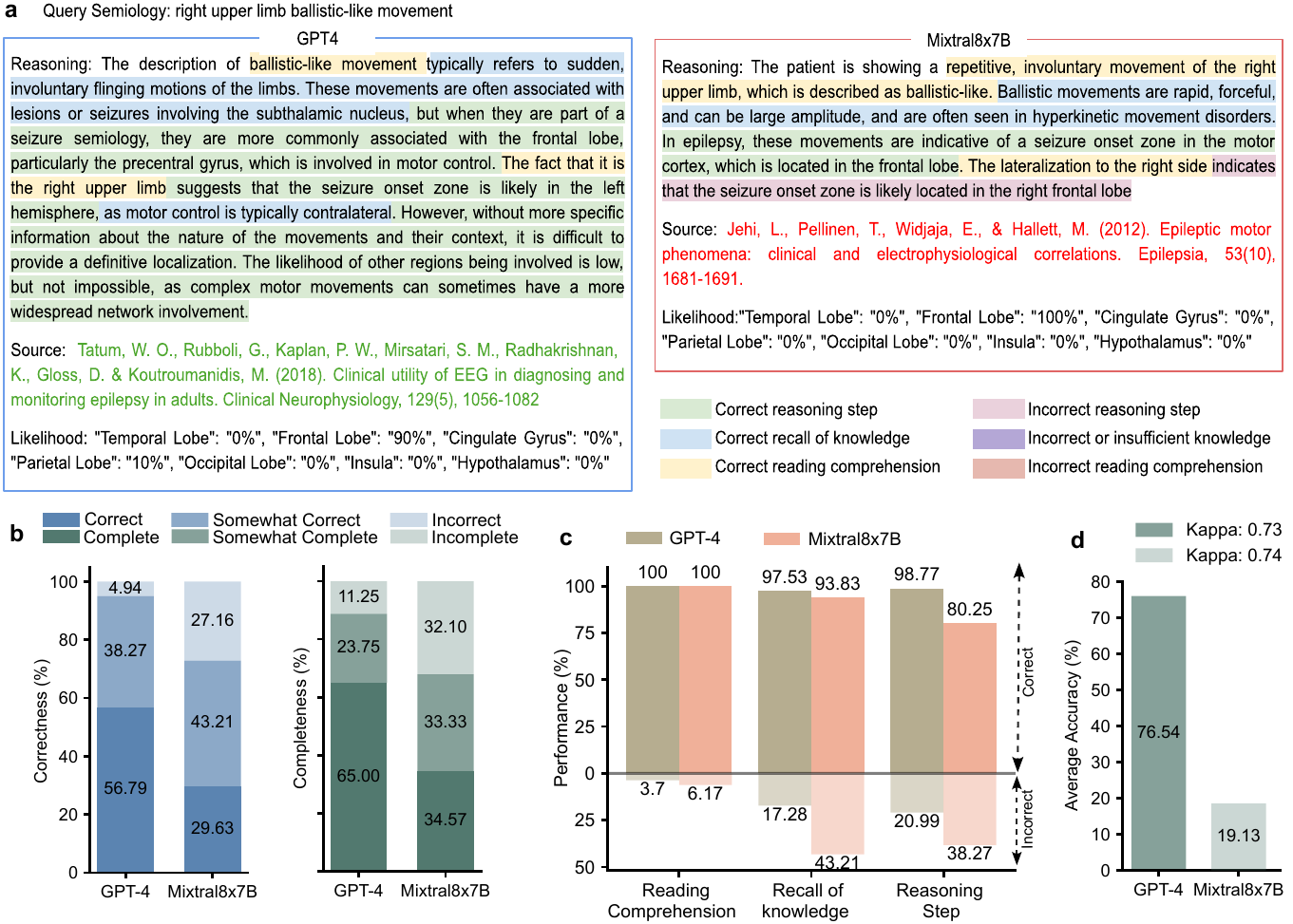}}
\caption{ Evaluation of model reasoning for representative random sample (n=81 queries) of dataset (a) Example query and corresponding annotations for a given semiology from GPT-4 and Mixtral-8x7B (b) Correctness and completeness of model outputs assessed by one clinician (c) Breakdown of model performance in reading comprehension, knowledge recall, and reasoning accuracy assessed by one clinician (d) Average citation accuracy across models assessed independently by two raters (inter-rater reliability: Cohen's kappa=0.73 for GPT-4, kappa=0.74 for Mixtral-8x7B). Bars represent proportion of correct responses out of 81 queries.
}
\label{fig:reasoning_res}
\vspace{-0.5cm}
\end{figure*}

\section{Results}
Our experimental pipeline of \emph{semioLLM} is illustrated in~\ref{fig:main}. Evaluated LLMs are given a text describing seizure symptoms taken from a public database ~\cite{semio2brain}. In the database, each seizure description is linked to one of 7 brain regions containing the respective seizure onset zone (SOZ). This ground truth was determined based on the fact that post-surgical resection of the SOZ patients remained seizure free for at least one year. 

\subsection{Prompt strategies significantly boost performance}
We assess classification performance for SOZ localization using the F1 score, comparing LLMs against a clinical evaluation and a naive classifier (lower bound, see Methods section~\ref{methods:baseline}). We further compare the zero shot setting (ZS) to 4 different prompt strategies: In the few-shot (FS)~\cite{brown2020language} condition, we leverage in-context learning (ICL) and provide each model with representative input-output pairs of 
 seizure descriptions and their SOZs. We employ Chain-of-thought (CoT) prompting~\cite{wei2022chain}, which provides step-by-step reasoning to improve the models’ ability to handle complex reasoning tasks. We also implement a task-specific strategy, FS-CoT, which combines the desired input-output mapping with a reasoning patterns curated by an expert clinician, mimicking the diagnostic reasoning of an epileptologist. Finally, we employ self-consistency (SC) ~\cite{wang2022self}, generating multiple reasoning paths and determining predictions through majority voting, which have been shown to enhance inference robustness and produce more reliable predictions for complex tasks.

As shown in~\ref{fig:corr_result}(a), most models perform just above chance level in the zero-shot (ZS) condition. Exceptions are Mixtral-8x7B and GPT-4, which achieved substantially higher F1 scores than all other models of 51.66\% and 52.27\%, respectively (95\% CI: [51.43, 51.90], [52.04, 52.50]), comparable to the clinician’s performance: Clinician 1 - 48.77\% (95\% CI: [48.53, 49.02]) and Clinician 2 - 46.75\% (95\% CI: [46.51, 46.99]). Importantly, we observed a substantial performance increase when introducing prompt-engineering across all models: Median F1 improvement relative to ZS was 6.49\% for Few-Shot prompting, 9.62\% for CoT, 9.49\% in FS-CoT, and 10.02\% increase for SC. Note that GPT-4, maintained a consistently high performance in all conditions,  only showing a modest gain from 52.27\% in ZS to 53.44\% with SC. In addition to general-purpose LLMs, we evaluated two medical-specific models—OpenBioLLM-70B and Med42-70B. OpenBioLLM-70B performed well, particularly in CoT (52.19) and SC (53.06),  approaching score in SC (53.44), but showed weaker ZS performance (38.56). Med42-70B exhibited better calibration than OpenBioLLM-70B (refer~\ref{fig:corr_result}(c)), yet lower overall scores (e.g., 46.70 in CoT, 47.63 in SC). While, both medical LLMs approached some general models in specific settings, neither consistently matched the performance of the top general-purpose models such as GPT-4.0 (mean 52.10) or Mixtral-8x7B (mean 50.58) - which achieved high accuracy for seizure-onset zone prediction, even without any external guidance (ZS). However, other models are able to match this through prompt-engineering. Interestingly, FS-CoT and SC both demonstrate the highest positive impact even though relying on different strategies-with FS-CoT providing expert-curated reasoning patterns that guide model outputs, and SC by enhancing robustness through the aggregation of multiple independent reasoning paths. 

\subsection{High confidence does not guarantee correctness}
To move towards trustworthy AI in risk-sensitive domains such as medicine, it is crucial to develop systems that are not only correct but also confident and well-calibrated in their outputs \cite{gal2016uncertainty, lyu2024calibrating}. Confidence, in practical terms, implies a higher degree of certainty in predictions. To assess this, we computed an entropy-based measure derived directly from model outputs rather than using a subjective self-assessment by LLMs~\cite{jungo2020analyzing, lyu2024calibrating, wimmer2023quantifying, zhou2020theory}. Specifically, we compute normalized Shannon's entropy ($H$)~\cite{shannon1948mathematical} using the likelihood estimates for different SOZs and derive a confidence score ($C = 1-H$) that ranges from 0 (lowest confidence) to 1 (highest confidence, see Eq.~\ref{eq:entropy} in Methods). 
Across all models, confidence scores were lowest in the zero-shot condition and improved consistently with prompt engineering (~\ref{fig:corr_result}(b)). For example, providing multiple solution examples in the few-shot condition increased the average model confidence by 13. 75\% compared to zero-shot, while task-specific reasoning crafted by an epileptologist in the FS-CoT condition resulted in a substantial 21\% increase in confidence, indicating that domain-specific demonstrations on task reasoning can significantly enhance the certainty of model predictions. Finally, as expected models exhibited the highest overall confidence for self-consistency (SC) with 35.25\% improvement as this prompt style is intended to reduce stochastic variability by aggregating predictions across multiple reasoning paths.
Assessing model confidence also requires evaluating calibration—that is how well model’s predicted probabilities align with its actual correctness. Even highly accurate models can produce probability estimates that do not reliably reflect their true likelihood of being correct~\cite{guo2017calibration}. Calibration can be quantified using the Brier score loss~\cite{brier1950verification, niculescu2005predicting}, where lower scores indicate better alignment.
As shown in ~\ref{fig:corr_result}(c), models exhibit a high variance in calibration in the zero-shot condition. This markedly improves for all models with more refined prompting techniques. Similar to the effect we observed when assessing model confidence through entropy, FS-CoT and SC are most efficient for aligning predicted probabilities more closely with actual accuracy. Notably, GPT-4 shows the best calibration, even in the zero-shot condition.  

\ref{fig:corr_result}(d) provides a summary of model performance across three key dimensions (average across prompt styles): accuracy (F1 score), confidence, and calibration. Larger enclosed areas indicate models that better balance these factors, highlighting trade-offs between predictive accuracy and reliability. Given the medical domain’s high-risk nature, the ideal model is one that not only achieves strong performance but also maintains well-calibrated confidence, reducing the likelihood of overly certain but incorrect decisions. Overall, two models - GPT-4 and Mixtral-8x7B - strike the best balance between all factors in our task. We therefore consider these two models for a more detailed investigation, including reasoning analysis and experimental manipulations such as in-context impersonation, narrative length effects, and cross-linguistic comparisons.

\subsection{Evaluating Clinical Reasoning and Source Attribution}
Thus far, our results demonstrate that LLMs can effectively map unstructured seizure descriptions to seizure onset zones (SOZs) in the brain. To explore how LLMs arrive at their decisions, we assessed the reasoning abilities of the two best performing models, GPT-4 and Mixtral-8x7B on a randomly selected subset of 81 chain-of-thought (CoT) responses using a clinical evaluation (see~\ref{extfig:userstudy_form} and Methods for details).
Following Med-PaLM~\cite{singhal2023large} and Liévin et al.'s~\cite{lievin2024can} protocol, we evaluated our models' reasoning outputs using three categories: correct/complete, somewhat correct/complete, and incorrect/incomplete. Additionally, each output is assessed for the proportion of correct and incorrect statements along three dimensions: (i) comprehension, (ii) knowledge recall, and (iii) logical reasoning. A representative example output, where both models correctly identify the seizure onset zone (SOZ), along with the annotations, is shown in ~\ref{fig:reasoning_res}(a). For the semiology ``right upper limb ballistic-like movement'', both models demonstrate accurate comprehension, knowledge recall and logical reasoning (yellow, blue and green). However, in contrast to GPT-4, Mixtral-8x7B misinterprets the associated hemisphere leading to an error (see pink text: ``right frontal lobe''). Additionally, Mixtral includes incorrect supporting scientific evidence, while GPT-4 cites a well-fitting and existing paper. These differences are well reflected in our summary analysis shown in~\ref{fig:reasoning_res}b-d. GPT-4 significantly outperformed Mixtral in both correctness (56.79\% vs. 29.63\%; z-test for proportions, p \textless 0.05) and completeness (65.00\% vs. 34.57\%; z-test for proportions, p \textless 0.05). Mixtral’s outputs were more often rated as ``somewhat correct'' and ``somewhat complete'' ((43.21\%,  33.33\% respectively). When examining specific dimensions (~\ref{fig:reasoning_res}(c)), GPT-4 made fewer errors for knowledge recall than Mixtral (17.28\% vs. 43.21\%; z-test for proportions, p \textless 0.05). However, the strongest differences were observed for logical reasoning: GPT-4 maintained a significantly higher score compared to Mixtral (98.77\% vs. 80.25\%; z-test for proportions, p \textless 0.05). Similarly, Mixtral's incorrect reasoning rate was almost twice that of GPT-4 (38.27\% vs. 20.99\%; z-test for proportions, p \textless 0.05). Additionally, \ref{fig:reasoning_res}(d) demonstrates GPT-4's superior citation accuracy (76.54\% for GPT-4 vs. 19.13\% for Mixtral), indicating GPT-4's enhanced ability to provide evidence-based sources for its decision. 

\begin{figure*}[]
\centering
\centerline{\includegraphics[width=1.0\textwidth]{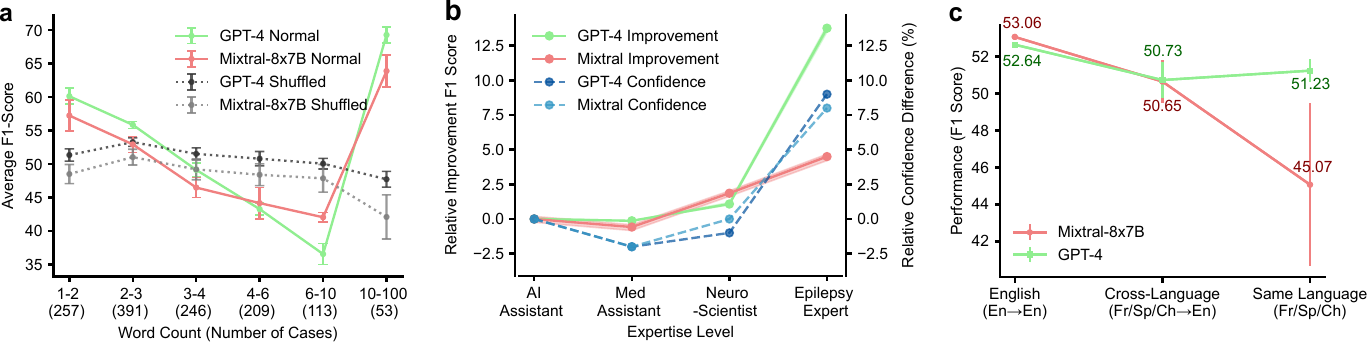}}
\caption{Impact of description length, persona adaptation and language on model performance (a) Performance across various description-length bins and length-shuffled inputs. Data show means $\pm$ SEM. Note that each range \([x,y)\) includes \(x\) but excludes \(y\) (b) Influence of in-context persona adaptation on zero-shot performance, shown as changes in F1 score (red/green) and confidence (blue/dark blue) relative to the AI assistant persona (n$=$ 1269 semiologies). (c) Effect of language variation on performance. In the ``English" condition, both the semiology description and prompt were in English. In the ``Cross-Language" condition, only the semiology description was in a different language. In the ``Same Language" condition, both the prompt and semiology description were in a non-English language. Data show means $\pm$ SD (n$=$ 1269 semiologies)
}
\label{fig:ablation}
\end{figure*}

\subsection{Factors influencing LLM performance in seizure diagnostics}

\subsubsection{Symptom description length}
Symptom descriptions vary in length—while longer descriptions may offer more information, they can also introduce irrelevant or contradictory details. To assess how this affects LLM-based diagnostic prediction, we categorized descriptions into six bins - based on word count and calculated the corresponding F1 scores, reporting the mean for each bin across all prompt strategies (refer ~\ref{fig:ablation}(a)). Interestingly, both models exhibit a distinctive U-shaped pattern, where predictions based on very short symptom descriptions initially achieve high performance, performance then decreases with increasing description length, and ultimately, observe the highest performance for the longest symptom descriptions. A Kruskal–Wallis test revealed statistically significant differences between the 
bins (p\textless 0.001 for both GPT-4 and Mixtral-8x7B).
 We validated this pattern against surrogate data obtained by randomly sampling semiologies from the original data, preserving the bin sizes. We then recomputed the F1 scores for these shuffled bins. Kolmogorov-Smirnov test (p\textless 0.001) between the original and shuffled data reveals that this U-shaped relationship represents a genuine effect of description length rather than a statistical artifact.

\subsubsection{Clinical in-context impersonation}
Clinical decision-making requires context-specific expertise. While prior work has shown that LLMs can improve performance by adopting expert personas in non-medical tasks \cite{salewski2024context}, we investigated whether this extends to clinical diagnostics. We prompted models to adopt increasingly specialized medical roles—medical assistant, neurologist, expert epileptologist—alongside a baseline AI assistant. Each prompt is prefixed with ``you are a \{persona\}'' and evaluated in a zero-shot setting to avoid few-shot example confounds.
For both GPT-4 and Mixtral-8x7B, performance, and confidence improved with increasing domain specificity, with the ``expert epileptologist'' yielding the highest scores (~\ref{fig:ablation}(b)). GPT-4 showed a substantial performance gain of 13.68\%, while Mixtral achieved a smaller but notable improvement (4.47\%). Confidence scores also increased with more specialized personas (GPT-4: 9\% and Mixtral: 8\%). These results demonstrate that in-context persona adaptation enhances both performance and confidence in clinical tasks, with GPT-4 more effectively leveraging contextual cues.

 \subsubsection{Multilingual Performance}
To assess the utility of LLMs for cross-lingual clinician-patient communication—particularly in medical tasks that depend entirely on verbal symptom descriptions—we compared performance across three language settings. In the baseline setting, both symptom descriptions and reasoning prompts were in English. In the mixed-language setting, symptom descriptions were in French, Spanish, or Chinese, while prompts remained in English. In the fully translated setting, both the clinical information and the reasoning instructions were presented entirely in French, Spanish, or Chinese, requiring full cross-lingual comprehension and reasoning. Our results (Figure~\ref{fig:ablation}(c)) show that both models perform best when prompt and input are in English (En→En), likely reflecting their English-centric training. Interestingly, in the cross-language setting (non-English symptom, English prompt), performance only drops slightly (GPT-4: –1.91\%, Mixtral: –2.41\%, n.s.). However, in the same-language setting (non-English symptoms and prompt), Mixtral's performance declines substantially by 8\%, while GPT-4 remains stable (–1.4\%, n.s.). This suggests both models can incorporate non-English input when anchored with English prompts, but especially Mixtral struggles in non-English contexts.

\section{Discussion}
Previous applications of LLMs in epilepsy have assessed general medical knowledge through structured Q\&A formats using single models such as ChatGPT~\cite{daungsupawong2024chatgpt, holgate2024extracting, kim2024assessing, luo2024clinical}, while NLP approaches in epilepsy have used rule-based or supervised models trained for narrow tasks like seizure type classification and frequency extraction~\cite{abeysinghe2025leveraging, decker2022development, mora2024nlp}. Most studies evaluated factual recall, rather than diagnostic reasoning and lacked grounding in real-world patient data. In contrast, our work \emph{SemioLLM} presents the first large-scale evaluation of 8 LLMs in diagnostic reasoning from over 1200 unstructured seizure descriptions. We observe that most LLMs can probabilistically infer seizure onset zones without structured input or domain-specific fine-tuning significantly above chance. Notably, GPT-4 and Mixtral-8x7B achieve performance comparable to a manual clinician-based assessment even under zero-shot conditions. Prompt engineering led to significant improvements in accuracy, confidence and calibration across all models - specifically clinician-guided chain-of-thought prompting led to the most substantial and consistent improvements. Our findings thus extend previous studies demonstrating the effectiveness of prompt engineering for structured medical Q\&A~\cite{singhal2023large}, clinical name entity recognition~\cite{hu2024improving}, and medical text summarization~\cite{sakai2025large} to probabilistic reasoning from free-text symptom descriptions and especially shows the potential for integrating clinician expertise into foundation model–based clinical decision systems~\cite{sonoda2025structured, wu2023large, zou2025rise}.

A key strength of LLMs is their capacity to generate explanatory reasoning alongside predictions—an important feature for transparency, interpretability, and trust in clinical decision support~\cite{ghassemi2021false}. Adapting established evaluation protocols from Med-PaLM~\cite{singhal2023large} and Liévin et al.\cite{lievin2024can}, we evaluated reasoning quality along with text comprehension, knowledge recall, and logical inference. Despite similar quantitative prediction performance, GPT-4's reasoning outputs were more often rated as correct and complete in contrast to Mixtral, with a particularly strong advantage in logical inference. In contrast, Mixtral made reasoning errors in over a third of its outputs. Differences in citation accuracy were particularly strong (GPT-4: 76.54\%, Mixtral: 19.13\%), reflecting ongoing challenges in factual grounding and source attribution for generative models~\cite{chen2023hallucinations, ji2023survey}.
These results underscore that performance metrics alone may obscure reasoning deficiencies, findings that are known from studies testing general medical knowledge \cite{singhal2023towards}. Here, retrieval-augmented generation (RAG)~\cite{lewis2020retrieval, mialon2023augmented} may help ground LLM reasoning in accurate, up-to-date knowledge, possibly improving reliability without retraining. 

Our study also reveals critical factors influencing LLM performance on this task. First, similar to prior results in non-medical domains~\cite{salewski2023context}, emulating an increasingly aligned clinical expert systematically improved both performance and confidence by 14\% and 10\% across models, respectively. Context-specific impersonation thus augments the ability of generalized language models to perform domain specific clinical tasks.
Second, prediction performance varied with symptom description length, showing accuracy differences of up to 32\%, where very short and highly detailed descriptions outperformed intermediate-length narratives. A possible explanation might be that brief descriptions closely match distinct canonical seizure features (e.g., “visual aura” for occipital onset~\cite{williamson1992occipital}), resembling the benefits of concise N-gram-based inputs in NLP tasks~\cite{mahendra2021impact}. Increasing description length, however, may introduce redundancies or contradicting evidence potentially degrading model performance ~\cite{liu2022note}. However, richly detailed but coherent input may offer enough structured context for LLMs to disambiguate and reason over complex clinical information. Ultimately these findings implicate that the level of detail as approximated through symptom description length might similarly influence diagnostic accuracy in LLMs and clinicians~\cite{muayqil2018accuracy}. 
Third, our evaluation showed that current LLMs can generalize across languages when anchored with English prompts, but also revealed limitations when all inputs are non-English. Both top-performing models, GPT-4 and Mixtral showed robust multilingual generalization when reasoning prompts were presented in English, even when symptom descriptions were provided in other languages such as French or Chinese, suggesting that they can integrate multilingual input when anchored with English-language instructions. In contrast, performance for one model (Mixtral) declined substantially when both prompts and input symptoms were non-English. This likely reflects the English-dominance of instruction tuning and pretraining corpora and points to limited cross-lingual generalization in current LLMs for the clinical domain, similar to limited multilingual performance in general-purpose NLP tasks ~\cite{hu2020xtreme}. Targeted multilingual instruction tuning, particularly in clinical domains, may thus be necessary not only to ensure robust and inclusive model behavior but also equitable application in multilingual healthcare systems. 

Unlike prior approaches that assess factual recall or classification, our framework combines quantitative metrics—correctness, confidence, and calibration — with qualitative expert annotation and evaluation of model reasoning. Importantly, LLMs can transform free-text clinical narratives into structured, actionable diagnostic inferences. Our work thus moves beyond basic knowledge verification and toward real-world clinical applicability. Importantly, our framework can be adapted to other medical domains. As a proof of concept, we provide example code and preliminary analyses in our GitHub repository in the domain of dermatology, linking skin anomaly descriptions to diagnostic classes. 

Despite these strengths, our study has several limitations. Our analysis is restricted to a single dataset which includes only adult focal epilepsy cases. While the diagnostic task (localization of SOZ) may apply to pediatric patients, it is not suitable for generic, generalized seizures. Additionally, while the dataset includes cases from diverse sources, we did not have the necessary metadata to analyze the impact of demographic and cultural variations in patient descriptions. While our translation-based cross-lingual analysis suggests that the language of seizure descriptions can affect performance, a key limitation is that our original corpus was monolingual. It will therefore be important in future studies to quantify language-specific effects in culturally diverse seizure descriptions. We demonstrate in our public code repository that the framework can be adapted to other clinical domains such as dermatology, our systematic evaluation and prompt optimization were conducted specifically for epilepsy. Consequently, prompt engineering strategies may need to be re-evaluated and tailored for other specialties before clinical deployment. Finally, as a comprehensive manual annotation would require impractical amounts of clinician effort, our reasoning analysis focused on a representative subset, chosen to match the overall distribution of the data. While this approach provides practical insight, it may introduce a degree of sampling bias or subjectivity, and future studies with larger-scale or multi-annotator reasoning evaluations would further strengthen the conclusions.

\section{Methods, Data and Code Availability}
Details on the SemioLLM methodology, including the full process for prompt structuring, dataset preparation, and model evaluation protocols, are provided in the Methods section available online after the references. Additionally, we include extended data analyses and visualizations to support the findings presented in the main text. 

The source code to reproduce the findings and figures are present at \href{https://github.com/liebelab/semiollm}{https://github.com/liebelab/semiollm} 

\section{Acknowledgments}
This work was supported by the Else Kröner Fresenius Foundation, Kolleg Clinbrain: Artificial Intelligence for Clinical Brain Research, the Clinician Scientist program of the Medical Faculty Tübingen funded by Deutsche Forschungsgemeinschaft (DFG, 493665037), the Priority Programme SPP 2241 - PN 520287829 (DFG), the Machine Learning Cluster of Excellence under Germany's Excellence Strategy – EXC number 2064/1 PN 390727645, the Tübingen AI Center, the Collaborative Research Center 1233 ``Robust Vision'' funded by the DFG and ERC (853489-DEXIM). M.D. is a member of the International Max Planck Research School for Intelligent Systems Tübingen (IMPRS-IS). The authors thank Prof. Dr. Jakob H. Macke and his lab for discussion and feedback; Almut Sophia Koepke, Shyamgopal Karthik, Matthias Tangemann and Elisa Nguyen for comments on the manuscript. 

\section{Author Contribution}
MD: Conceptualization, Methodology, Validation, Formal Analysis, Data Curation, Reasoning Study Design and Analysis, Visualization, Writing - Original Draft, Writing - Review and Editing. 
MJP: Methodology (design of clinical study form); Formal Analysis (compilation and evaluation of clinical responses and model source citations), Writing - Review and Editing. 
FP: Clinical evaluation, Writing - Review.
ZA: Supervision, Funding Acquisition.
SL: Conceptualization, Methodology, Clinical evaluation, Project Administration, Supervision, Funding Acquisition, Writing - Review and Editing.

\bibliography{egbib}

\begin{thebibliography}{71}
\providecommand{\natexlab}[1]{#1}
\providecommand{\url}[1]{\texttt{#1}}
\expandafter\ifx\csname urlstyle\endcsname\relax
  \providecommand{\doi}[1]{doi: #1}\else
  \providecommand{\doi}{doi: \begingroup \urlstyle{rm}\Url}\fi

\bibitem[Abeysinghe et~al.(2025)Abeysinghe, Tao, Lhatoo, Zhang, and Cui]{abeysinghe2025leveraging}
Abeysinghe, R., Tao, S., Lhatoo, S.~D., Zhang, G.-Q., and Cui, L.
\newblock Leveraging pretrained language models for seizure frequency extraction from epilepsy evaluation reports.
\newblock \emph{npj Digital Medicine}, 8\penalty0 (1):\penalty0 208, 2025.

\bibitem[Achiam et~al.(2023)Achiam, Adler, Agarwal, Ahmad, Akkaya, Aleman, Almeida, Altenschmidt, Altman, Anadkat, et~al.]{achiam2023gpt4}
Achiam, J., Adler, S., Agarwal, S., Ahmad, L., Akkaya, I., Aleman, F.~L., Almeida, D., Altenschmidt, J., Altman, S., Anadkat, S., et~al.
\newblock Gpt-4 technical report.
\newblock \emph{arXiv preprint arXiv:2303.08774}, 2023.

\bibitem[Alaa et~al.(2025)Alaa, Hartvigsen, Golchini, Dutta, Dean, Raji, and Zack]{alaa2025medical}
Alaa, A., Hartvigsen, T., Golchini, N., Dutta, S., Dean, F., Raji, I.~D., and Zack, T.
\newblock Medical large language model benchmarks should prioritize construct validity.
\newblock \emph{arXiv preprint arXiv:2503.10694}, 2025.

\bibitem[Alim-Marvasti et~al.(2022)Alim-Marvasti, Romagnoli, et~al.]{semio2brain}
Alim-Marvasti, A., Romagnoli, et~al.
\newblock Probabilistic landscape of seizure semiology localizing values.
\newblock \emph{Brain Communications}, 4\penalty0 (3):\penalty0 fcac130, 2022.

\bibitem[Bai et~al.(2023)Bai, Bai, Chu, Cui, Dang, Deng, Fan, Ge, Han, Huang, et~al.]{bai2023qwen}
Bai, J., Bai, S., Chu, Y., Cui, Z., Dang, K., Deng, X., Fan, Y., Ge, W., Han, Y., Huang, F., et~al.
\newblock Qwen technical report.
\newblock \emph{arXiv preprint arXiv:2309.16609}, 2023.

\bibitem[Beniczky et~al.(2022)Beniczky, Tatum, Blumenfeld, Stefan, Mani, Maillard, Fahoum, Vinayan, Mayor, Vlachou, et~al.]{beniczky2022seizureilae}
Beniczky, S., Tatum, W.~O., Blumenfeld, H., Stefan, H., Mani, J., Maillard, L., Fahoum, F., Vinayan, K.~P., Mayor, L.~C., Vlachou, M., et~al.
\newblock Seizure semiology: Ilae glossary of terms and their significance.
\newblock \emph{Epileptic Disorders}, 24\penalty0 (3):\penalty0 447--495, 2022.

\bibitem[Bolton et~al.(2024)Bolton, Venigalla, Yasunaga, Hall, Xiong, Lee, Daneshjou, Frankle, Liang, Carbin, et~al.]{bolton2024biomedlm}
Bolton, E., Venigalla, A., Yasunaga, M., Hall, D., Xiong, B., Lee, T., Daneshjou, R., Frankle, J., Liang, P., Carbin, M., et~al.
\newblock Biomedlm: A 2.7 b parameter language model trained on biomedical text.
\newblock \emph{arXiv preprint arXiv:2403.18421}, 2024.

\bibitem[Bommasani et~al.(2021)Bommasani, Hudson, Adeli, Altman, Arora, von Arx, Bernstein, Bohg, Bosselut, Brunskill, et~al.]{bommasani2021opportunities}
Bommasani, R., Hudson, D.~A., Adeli, E., Altman, R., Arora, S., von Arx, M., Bernstein, M., Bohg, J., Bosselut, A., Brunskill, E., et~al.
\newblock On the opportunities and risks of foundation models.
\newblock \emph{arXiv preprint arXiv:2108.07258}, 2021.

\bibitem[Brier(1950)]{brier1950verification}
Brier, G.~W.
\newblock Verification of forecasts expressed in terms of probability.
\newblock \emph{Monthly weather review}, 78\penalty0 (1):\penalty0 1--3, 1950.

\bibitem[Brown et~al.(2020)Brown, Mann, Ryder, Subbiah, Kaplan, Dhariwal, Neelakantan, Shyam, Sastry, Askell, et~al.]{brown2020language}
Brown, T., Mann, B., Ryder, N., Subbiah, M., Kaplan, J.~D., Dhariwal, P., Neelakantan, A., Shyam, P., Sastry, G., Askell, A., et~al.
\newblock Language models are few-shot learners.
\newblock \emph{Advances in neural information processing systems}, 33:\penalty0 1877--1901, 2020.

\bibitem[Chen et~al.(2023{\natexlab{a}})Chen, Tian, Xiong, Leng, Zubeldía, Gan, Xu, Yue, Gong, Bai, et~al.]{chen2023hallucinations}
Chen, Y., Tian, S., Xiong, H., Leng, Y., Zubeldía, A.~R., Gan, K., Xu, J., Yue, H., Gong, M., Bai, S., et~al.
\newblock Hallucinations in large language models in healthcare: a survey of scenarios, factors, mitigations, and research directions.
\newblock \emph{arXiv preprint arXiv:2311.05232}, 2023{\natexlab{a}}.

\bibitem[Chen et~al.(2023{\natexlab{b}})Chen, Cano, Romanou, Bonnet, Matoba, Salvi, Pagliardini, Fan, K{\"o}pf, Mohtashami, et~al.]{chen2023meditron}
Chen, Z., Cano, A.~H., Romanou, A., Bonnet, A., Matoba, K., Salvi, F., Pagliardini, M., Fan, S., K{\"o}pf, A., Mohtashami, A., et~al.
\newblock Meditron-70b: Scaling medical pretraining for large language models.
\newblock \emph{arXiv preprint arXiv:2311.16079}, 2023{\natexlab{b}}.

\bibitem[Christophe et~al.(2024)Christophe, Kanithi, Raha, Khan, and Pimentel]{christophe2024med42}
Christophe, C., Kanithi, P.~K., Raha, T., Khan, S., and Pimentel, M.~A.
\newblock Med42-v2: A suite of clinical llms.
\newblock \emph{arXiv preprint arXiv:2408.06142}, 2024.

\bibitem[Daungsupawong \& Wiwanitkit(2024)Daungsupawong and Wiwanitkit]{daungsupawong2024chatgpt}
Daungsupawong, H. and Wiwanitkit, V.
\newblock Chatgpt's responses to questions related to epilepsy.
\newblock \emph{Seizure-European Journal of Epilepsy}, 114:\penalty0 105, 2024.

\bibitem[Decker et~al.(2022)Decker, Turco, Xu, Terman, Kosaraju, Jamil, Davis, Litt, Ellis, Khankhanian, et~al.]{decker2022development}
Decker, B.~M., Turco, A., Xu, J., Terman, S.~W., Kosaraju, N., Jamil, A., Davis, K.~A., Litt, B., Ellis, C.~A., Khankhanian, P., et~al.
\newblock Development of a natural language processing algorithm to extract seizure types and frequencies from the electronic health record.
\newblock \emph{Seizure: European Journal of Epilepsy}, 101:\penalty0 48--51, 2022.

\bibitem[Devlin et~al.(2019)Devlin, Chang, Lee, and Toutanova]{devlin2019bert}
Devlin, J., Chang, M.-W., Lee, K., and Toutanova, K.
\newblock Bert: Pre-training of deep bidirectional transformers for language understanding.
\newblock In \emph{Proceedings of the 2019 conference of the North American chapter of the association for computational linguistics: human language technologies, volume 1 (long and short papers)}, pp.\  4171--4186, 2019.

\bibitem[Dong et~al.(2022)Dong, Li, Dai, Zheng, Wu, Chang, Sun, Xu, and Sui]{dong2022survey}
Dong, Q., Li, L., Dai, D., Zheng, C., Wu, Z., Chang, B., Sun, X., Xu, J., and Sui, Z.
\newblock A survey on in-context learning.
\newblock \emph{arXiv preprint arXiv:2301.00234}, 2022.

\bibitem[Dubey et~al.(2024)Dubey, Jauhri, Pandey, Kadian, Al-Dahle, Letman, Mathur, Schelten, Yang, Fan, et~al.]{dubey2024llama3}
Dubey, A., Jauhri, A., Pandey, A., Kadian, A., Al-Dahle, A., Letman, A., Mathur, A., Schelten, A., Yang, A., Fan, A., et~al.
\newblock The llama 3 herd of models.
\newblock \emph{arXiv preprint arXiv:2407.21783}, 2024.

\bibitem[ESR(2023)]{european2023esr}
ESR.
\newblock Esr paper on structured reporting in radiology—update 2023.
\newblock \emph{Insights into Imaging}, 14\penalty0 (1):\penalty0 199, 2023.

\bibitem[Gal et~al.(2016)]{gal2016uncertainty}
Gal, Y. et~al.
\newblock Uncertainty in deep learning.
\newblock 2016.

\bibitem[Ghassemi et~al.(2021)Ghassemi, Oakden-Rayner, and Beam]{ghassemi2021false}
Ghassemi, M., Oakden-Rayner, L., and Beam, A.~L.
\newblock The false hope of current approaches to explainable artificial intelligence in health care.
\newblock \emph{The Lancet Digital Health}, 3\penalty0 (11):\penalty0 e745--e750, 2021.

\bibitem[Guo et~al.(2017)Guo, Pleiss, Sun, and Weinberger]{guo2017calibration}
Guo, C., Pleiss, G., Sun, Y., and Weinberger, K.~Q.
\newblock On calibration of modern neural networks.
\newblock In \emph{Proceedings of the 34th International Conference on Machine Learning (ICML)}, volume~70, pp.\  1321--1330. PMLR, 2017.

\bibitem[Hager et~al.(2024)Hager, Jungmann, Holland, Bhagat, Hubrecht, Knauer, Vielhauer, Makowski, Braren, Kaissis, et~al.]{hager2024evaluation}
Hager, P., Jungmann, F., Holland, R., Bhagat, K., Hubrecht, I., Knauer, M., Vielhauer, J., Makowski, M., Braren, R., Kaissis, G., et~al.
\newblock Evaluation and mitigation of the limitations of large language models in clinical decision-making.
\newblock \emph{Nature medicine}, 30\penalty0 (9):\penalty0 2613--2622, 2024.

\bibitem[Holgate et~al.(2024)Holgate, Fang, Shek, McWilliam, Viana, Winston, Teo, and Richardson]{holgate2024extracting}
Holgate, B., Fang, S., Shek, A., McWilliam, M., Viana, P., Winston, J.~S., Teo, J.~T., and Richardson, M.~P.
\newblock Extracting epilepsy patient data with llama 2.
\newblock In \emph{Proceedings of the 23rd Workshop on Biomedical Natural Language Processing}, pp.\  526--535, 2024.

\bibitem[Hu et~al.(2020)Hu, Ruder, Siddhant, Neubig, Firat, and Johnson]{hu2020xtreme}
Hu, J., Ruder, S., Siddhant, A., Neubig, G., Firat, O., and Johnson, M.
\newblock Xtreme: A massively multilingual multi-task benchmark for evaluating cross-lingual generalisation.
\newblock In \emph{International conference on machine learning}, pp.\  4411--4421. PMLR, 2020.

\bibitem[Hu et~al.(2024)Hu, Chen, Du, Peng, Keloth, Zuo, Zhou, Li, Jiang, Lu, et~al.]{hu2024improving}
Hu, Y., Chen, Q., Du, J., Peng, X., Keloth, V.~K., Zuo, X., Zhou, Y., Li, Z., Jiang, X., Lu, Z., et~al.
\newblock Improving large language models for clinical named entity recognition via prompt engineering.
\newblock \emph{Journal of the American Medical Informatics Association}, 31\penalty0 (9):\penalty0 1812--1820, 2024.

\bibitem[Huang et~al.(2019)Huang, Altosaar, and Ranganath]{huang2019clinicalbert}
Huang, K., Altosaar, J., and Ranganath, R.
\newblock Clinicalbert: Modeling clinical notes and predicting hospital readmission.
\newblock \emph{arXiv preprint arXiv:1904.05342}, 2019.

\bibitem[Ji et~al.(2023)Ji, Lee, Frieske, Yu, Su, Xu, Ishii, Bang, Dai, Madotto, et~al.]{ji2023survey}
Ji, Z., Lee, N., Frieske, R., Yu, T., Su, D., Xu, Y., Ishii, E., Bang, Y.~J., Dai, A., Madotto, A., et~al.
\newblock A survey of hallucination in large language models.
\newblock \emph{ACM Computing Surveys}, 2023.

\bibitem[Jiang et~al.(2024)Jiang, Sablayrolles, Roux, Mensch, Savary, Bamford, Chaplot, Casas, Hanna, Bressand, et~al.]{jiang2024mixtral}
Jiang, A.~Q., Sablayrolles, A., Roux, A., Mensch, A., Savary, B., Bamford, C., Chaplot, D.~S., Casas, D. d.~l., Hanna, E.~B., Bressand, F., et~al.
\newblock Mixtral of experts.
\newblock \emph{arXiv preprint arXiv:2401.04088}, 2024.

\bibitem[Jin et~al.(2021)Jin, Pan, Oufattole, Weng, Fang, and Szolovits]{jin2021medqa}
Jin, D., Pan, E., Oufattole, N., Weng, W.-H., Fang, H., and Szolovits, P.
\newblock What disease does this patient have? a large-scale open domain question answering dataset from medical exams.
\newblock \emph{Applied Sciences}, 11\penalty0 (14):\penalty0 6421, 2021.

\bibitem[Jin et~al.(2019)Jin, Dhingra, Liu, Cohen, and Lu]{jin2019pubmedqa}
Jin, Q., Dhingra, B., Liu, Z., Cohen, W., and Lu, X.
\newblock Pubmedqa: A dataset for biomedical research question answering.
\newblock In \emph{Proceedings of the 2019 Conference on Empirical Methods in Natural Language Processing and the 9th International Joint Conference on Natural Language Processing (EMNLP-IJCNLP)}, pp.\  2567--2577, 2019.

\bibitem[Jungo et~al.(2020)Jungo, Scheidegger, and Reyes]{jungo2020analyzing}
Jungo, A., Scheidegger, O., and Reyes, M.
\newblock Analyzing the quality and challenges of uncertainty quantification in medical image segmentation.
\newblock \emph{Medical Image Analysis}, 64:\penalty0 101723, 2020.

\bibitem[Kim et~al.(2024)Kim, Shin, Kim, Lee, and Cho]{kim2024assessing}
Kim, H.-W., Shin, D.-H., Kim, J., Lee, G.-H., and Cho, J.~W.
\newblock Assessing the performance of chatgpt's responses to questions related to epilepsy: a cross-sectional study on natural language processing and medical information retrieval.
\newblock \emph{Seizure: European Journal of Epilepsy}, 114:\penalty0 1--8, 2024.

\bibitem[Krithara et~al.(2023)Krithara, Nentidis, Bougiatiotis, and Paliouras]{krithara2023bioasq}
Krithara, A., Nentidis, A., Bougiatiotis, K., and Paliouras, G.
\newblock Bioasq-qa: A manually curated corpus for biomedical question answering.
\newblock \emph{Scientific Data}, 10\penalty0 (1):\penalty0 170, 2023.

\bibitem[Lewis et~al.(2020)Lewis, Perez, Piktus, Petroni, Karpukhin, Goyal, K{\"u}ttler, Lewis, Yih, Rockt{\"a}schel, et~al.]{lewis2020retrieval}
Lewis, P., Perez, E., Piktus, A., Petroni, F., Karpukhin, V., Goyal, N., K{\"u}ttler, H., Lewis, M., Yih, W.-t., Rockt{\"a}schel, T., et~al.
\newblock Retrieval-augmented generation for knowledge-intensive nlp tasks.
\newblock \emph{Advances in neural information processing systems}, 33:\penalty0 9459--9474, 2020.

\bibitem[Li{\'e}vin et~al.(2024)Li{\'e}vin, Hother, Motzfeldt, and Winther]{lievin2024can}
Li{\'e}vin, V., Hother, C.~E., Motzfeldt, A.~G., and Winther, O.
\newblock Can large language models reason about medical questions?
\newblock \emph{Patterns}, 5\penalty0 (3), 2024.

\bibitem[Liu et~al.(2022)Liu, Capurro, Nguyen, and Verspoor]{liu2022note}
Liu, J., Capurro, D., Nguyen, A., and Verspoor, K.
\newblock “note bloat” impacts deep learning-based nlp models for clinical prediction tasks.
\newblock \emph{Journal of biomedical informatics}, 133:\penalty0 104149, 2022.

\bibitem[L{\"u}ders et~al.(2006)L{\"u}ders, Najm, Nair, Widdess-Walsh, and Bingman]{luders2006epileptogenic}
L{\"u}ders, H.~O., Najm, I., Nair, D., Widdess-Walsh, P., and Bingman, W.
\newblock The epileptogenic zone: general principles.
\newblock \emph{Epileptic disorders}, 8:\penalty0 S1--S9, 2006.

\bibitem[Luo et~al.(2024)Luo, Jiao, Fotedar, Ding, Karakis, Rao, Asmar, Xian, Aboud, Wen, et~al.]{luo2024clinical}
Luo, Y., Jiao, M., Fotedar, N., Ding, J.-E., Karakis, I., Rao, V.~R., Asmar, M., Xian, X., Aboud, O., Wen, Y., et~al.
\newblock The clinical value of chatgpt for epilepsy presurgical decision making: Systematic evaluation on seizure semiology interpretation.
\newblock \emph{medRxiv}, pp.\  2024--04, 2024.

\bibitem[Lyu et~al.(2024)Lyu, Shridhar, Malaviya, Zhang, Elazar, Tandon, Apidianaki, Sachan, and Callison-Burch]{lyu2024calibrating}
Lyu, Q., Shridhar, K., Malaviya, C., Zhang, L., Elazar, Y., Tandon, N., Apidianaki, M., Sachan, M., and Callison-Burch, C.
\newblock Calibrating large language models with sample consistency.
\newblock \emph{arXiv preprint arXiv:2402.13904}, 2024.

\bibitem[Mahendra et~al.(2021)Mahendra, Luo, Mills, Schenk, Butte, and Dudley]{mahendra2021impact}
Mahendra, M., Luo, Y., Mills, H., Schenk, G., Butte, A.~J., and Dudley, R.~A.
\newblock Impact of different approaches to preparing notes for analysis with natural language processing on the performance of prediction models in intensive care.
\newblock \emph{Critical care explorations}, 3\penalty0 (6):\penalty0 e0450, 2021.

\bibitem[Mialon et~al.(2023)Mialon, Dessì, Lomeli, Nalmpantis, Pasunuru, Raileanu, Rozière, Schick, Dwivedi-Yu, Celikyilmaz, et~al.]{mialon2023augmented}
Mialon, G., Dessì, R., Lomeli, M., Nalmpantis, C., Pasunuru, R., Raileanu, R., Rozière, B., Schick, T., Dwivedi-Yu, J., Celikyilmaz, A., et~al.
\newblock Augmented language models: a survey.
\newblock \emph{arXiv preprint arXiv:2302.07842}, 2023.

\bibitem[Mora et~al.(2024)Mora, Turrisi, Chiarella, Consales, Tassi, Mai, Nobili, Barla, and Arnulfo]{mora2024nlp}
Mora, S., Turrisi, R., Chiarella, L., Consales, A., Tassi, L., Mai, R., Nobili, L., Barla, A., and Arnulfo, G.
\newblock Nlp-based tools for localization of the epileptogenic zone in patients with drug-resistant focal epilepsy.
\newblock \emph{Scientific Reports}, 14\penalty0 (1):\penalty0 2349, 2024.

\bibitem[Muayqil et~al.(2018)Muayqil, Alanazy, Almalak, Alsalman, Abdulfattah, Aldraihem, Al-Hussain, and Aljafen]{muayqil2018accuracy}
Muayqil, T.~A., Alanazy, M.~H., Almalak, H.~M., Alsalman, H.~K., Abdulfattah, F.~W., Aldraihem, A.~I., Al-Hussain, F., and Aljafen, B.~N.
\newblock Accuracy of seizure semiology obtained from first-time seizure witnesses.
\newblock \emph{BMC neurology}, 18:\penalty0 1--6, 2018.

\bibitem[Niculescu-Mizil \& Caruana(2005)Niculescu-Mizil and Caruana]{niculescu2005predicting}
Niculescu-Mizil, A. and Caruana, R.
\newblock Predicting good probabilities with supervised learning.
\newblock In \emph{Proceedings of the 22nd international conference on Machine learning}, pp.\  625--632, 2005.

\bibitem[Nowacki \& Jirsch(2017)Nowacki and Jirsch]{nowacki2017evaluation}
Nowacki, T.~A. and Jirsch, J.~D.
\newblock Evaluation of the first seizure patient: Key points in the history and physical examination.
\newblock \emph{Seizure}, 49:\penalty0 54--63, 2017.

\bibitem[Page et~al.(2021)Page, McKenzie, Bossuyt, Boutron, Hoffmann, Mulrow, Shamseer, Tetzlaff, Akl, Brennan, et~al.]{page2021prisma}
Page, M.~J., McKenzie, J.~E., Bossuyt, P.~M., Boutron, I., Hoffmann, T.~C., Mulrow, C.~D., Shamseer, L., Tetzlaff, J.~M., Akl, E.~A., Brennan, S.~E., et~al.
\newblock The prisma 2020 statement: an updated guideline for reporting systematic reviews.
\newblock \emph{International journal of surgery}, 88:\penalty0 105906, 2021.

\bibitem[Pal et~al.(2022)Pal, Umapathi, and Sankarasubbu]{pal2022medmcqa}
Pal, A., Umapathi, L.~K., and Sankarasubbu, M.
\newblock Medmcqa: A large-scale multi-subject multi-choice dataset for medical domain question answering.
\newblock In \emph{Conference on health, inference, and learning}, pp.\  248--260. PMLR, 2022.

\bibitem[Raji et~al.(2025)Raji, Daneshjou, and Alsentzer]{raji2025s}
Raji, I.~D., Daneshjou, R., and Alsentzer, E.
\newblock It’s time to bench the medical exam benchmark, 2025.

\bibitem[Sakai \& Lam(2025)Sakai and Lam]{sakai2025large}
Sakai, H. and Lam, S.~S.
\newblock Large language models for healthcare text classification: A systematic review.
\newblock \emph{arXiv preprint arXiv:2503.01159}, 2025.

\bibitem[Salewski et~al.(2023)Salewski, Alaniz, Rio-Torto, Schulz, and Akata]{salewski2023context}
Salewski, L., Alaniz, S., Rio-Torto, I., Schulz, E., and Akata, Z.
\newblock In-context impersonation reveals large language models' strengths and biases.
\newblock \emph{Advances in neural information processing systems}, 36:\penalty0 72044--72057, 2023.

\bibitem[Salewski et~al.(2024)Salewski, Alaniz, Rio-Torto, Schulz, and Akata]{salewski2024context}
Salewski, L., Alaniz, S., Rio-Torto, I., Schulz, E., and Akata, Z.
\newblock In-context impersonation reveals large language models' strengths and biases.
\newblock \emph{Advances in Neural Information Processing Systems}, 36, 2024.

\bibitem[Sarvari et~al.(2024)Sarvari, Al-fagih, Ghuwel, and Al-fagih]{sarvari2024systematic}
Sarvari, P., Al-fagih, Z., Ghuwel, A., and Al-fagih, O.
\newblock A systematic evaluation of the performance of gpt-4 and palm2 to diagnose comorbidities in mimic-iv patients.
\newblock \emph{Health Care Science}, 2024.

\bibitem[Savage et~al.(2024)Savage, Nayak, Gallo, Rangan, and Chen]{savage2024diagnostic}
Savage, T., Nayak, A., Gallo, R., Rangan, E., and Chen, J.~H.
\newblock Diagnostic reasoning prompts reveal the potential for large language model interpretability in medicine.
\newblock \emph{NPJ Digital Medicine}, 7\penalty0 (1):\penalty0 20, 2024.

\bibitem[Shannon(1948)]{shannon1948mathematical}
Shannon, C.~E.
\newblock A mathematical theory of communication.
\newblock \emph{The Bell system technical journal}, 27\penalty0 (3):\penalty0 379--423, 1948.

\bibitem[Singhal et~al.(2023{\natexlab{a}})Singhal, Azizi, Tu, Mahdavi, Wei, Chung, Scales, Tanwani, Cole-Lewis, Pfohl, et~al.]{singhal2023large}
Singhal, K., Azizi, S., Tu, T., Mahdavi, S.~S., Wei, J., Chung, H.~W., Scales, N., Tanwani, A., Cole-Lewis, H., Pfohl, S., et~al.
\newblock Large language models encode clinical knowledge.
\newblock \emph{Nature}, 620\penalty0 (7972):\penalty0 172--180, 2023{\natexlab{a}}.

\bibitem[Singhal et~al.(2023{\natexlab{b}})Singhal, Tu, Gottweis, Sayres, Wulczyn, Hou, Clark, Pfohl, Cole-Lewis, Neal, et~al.]{singhal2023towards}
Singhal, K., Tu, T., Gottweis, J., Sayres, R., Wulczyn, E., Hou, L., Clark, K., Pfohl, S., Cole-Lewis, H., Neal, D., et~al.
\newblock Towards expert-level medical question answering with large language models.
\newblock \emph{arXiv preprint arXiv:2305.09617}, 2023{\natexlab{b}}.

\bibitem[Sisodiya \& Goldstein(2007)Sisodiya and Goldstein]{sisodiya2007dre}
Sisodiya, S.~M. and Goldstein, D.~B.
\newblock Drug resistance in epilepsy: more twists in the tale.
\newblock \emph{Epilepsia}, 48\penalty0 (12):\penalty0 2369--2370, 2007.

\bibitem[Sonoda et~al.(2025)Sonoda, Kurokawa, Hagiwara, Asari, Fukushima, Kanzawa, Gonoi, and Abe]{sonoda2025structured}
Sonoda, Y., Kurokawa, R., Hagiwara, A., Asari, Y., Fukushima, T., Kanzawa, J., Gonoi, W., and Abe, O.
\newblock Structured clinical reasoning prompt enhances llm’s diagnostic capabilities in diagnosis please quiz cases.
\newblock \emph{Japanese Journal of Radiology}, 43\penalty0 (4):\penalty0 586--592, 2025.

\bibitem[Touvron et~al.(2023)Touvron, Martin, Stone, Albert, Almahairi, Babaei, Bashlykov, Batra, Bhargava, Bhosale, et~al.]{touvron2023llama2}
Touvron, H., Martin, L., Stone, K., Albert, P., Almahairi, A., Babaei, Y., Bashlykov, N., Batra, S., Bhargava, P., Bhosale, S., et~al.
\newblock Llama 2: Open foundation and fine-tuned chat models.
\newblock \emph{arXiv preprint arXiv:2307.09288}, 2023.

\bibitem[Tufenkjian \& L{\"u}ders(2012)Tufenkjian and L{\"u}ders]{tufenkjian2012seizure}
Tufenkjian, K. and L{\"u}ders, H.~O.
\newblock Seizure semiology: its value and limitations in localizing the epileptogenic zone.
\newblock \emph{Journal of clinical neurology (Seoul, Korea)}, 8\penalty0 (4):\penalty0 243, 2012.

\bibitem[van Diessen et~al.(2024)van Diessen, van Amerongen, Zijlmans, and Otte]{van2024potential}
van Diessen, E., van Amerongen, R.~A., Zijlmans, M., and Otte, W.~M.
\newblock Potential merits and flaws of large language models in epilepsy care: A critical review.
\newblock \emph{Epilepsia}, 2024.

\bibitem[Wang et~al.(2022)Wang, Wei, Schuurmans, Le, Chi, Narang, Chowdhery, and Zhou]{wang2022self}
Wang, X., Wei, J., Schuurmans, D., Le, Q., Chi, E., Narang, S., Chowdhery, A., and Zhou, D.
\newblock Self-consistency improves chain of thought reasoning in language models.
\newblock \emph{arXiv preprint arXiv:2203.11171}, 2022.

\bibitem[Wei et~al.(2022)Wei, Wang, Schuurmans, Bosma, Xia, Chi, Le, Zhou, et~al.]{wei2022chain}
Wei, J., Wang, X., Schuurmans, D., Bosma, M., Xia, F., Chi, E., Le, Q.~V., Zhou, D., et~al.
\newblock Chain-of-thought prompting elicits reasoning in large language models.
\newblock \emph{Advances in neural information processing systems}, 35:\penalty0 24824--24837, 2022.

\bibitem[Wiebe et~al.(2001)Wiebe, Blume, Girvin, and Eliasziw]{wiebe2001randomizednejm}
Wiebe, S., Blume, W.~T., Girvin, J.~P., and Eliasziw, M.
\newblock A randomized, controlled trial of surgery for temporal-lobe epilepsy.
\newblock \emph{New England Journal of Medicine}, 345\penalty0 (5):\penalty0 311--318, 2001.

\bibitem[Williamson et~al.(1992)Williamson, Thadani, Darcey, Spencer, Spencer, and Mattson]{williamson1992occipital}
Williamson, P., Thadani, V., Darcey, T., Spencer, D., Spencer, S., and Mattson, R.
\newblock Occipital lobe epilepsy: clinical characteristics, seizure spread patterns, and results of surgery.
\newblock \emph{Annals of Neurology: Official Journal of the American Neurological Association and the Child Neurology Society}, 31\penalty0 (1):\penalty0 3--13, 1992.

\bibitem[Wimmer et~al.(2023)Wimmer, Sale, Hofman, Bischl, and H{\"u}llermeier]{wimmer2023quantifying}
Wimmer, L., Sale, Y., Hofman, P., Bischl, B., and H{\"u}llermeier, E.
\newblock Quantifying aleatoric and epistemic uncertainty in machine learning: Are conditional entropy and mutual information appropriate measures?
\newblock In \emph{Uncertainty in Artificial Intelligence}, pp.\  2282--2292. PMLR, 2023.

\bibitem[Wu et~al.(2023)Wu, Chen, and Chen]{wu2023large}
Wu, C.-K., Chen, W.-L., and Chen, H.-H.
\newblock Large language models perform diagnostic reasoning.
\newblock \emph{arXiv preprint arXiv:2307.08922}, 2023.

\bibitem[Yan et~al.(2024)Yan, Li, Zhang, Yin, Fei, Peng, Bi, Feng, Chen, Liu, et~al.]{yan2024large}
Yan, L.~K., Li, M., Zhang, Y., Yin, C.~H., Fei, C., Peng, B., Bi, Z., Feng, P., Chen, K., Liu, J., et~al.
\newblock Large language model benchmarks in medical tasks.
\newblock \emph{arXiv preprint arXiv:2410.21348}, 2024.

\bibitem[Zhou(2020)]{zhou2020theory}
Zhou, M.
\newblock A theory on ai uncertainty based on rademacher complexity and shannon entropy.
\newblock In \emph{2020 IEEE 3rd International Conference of Safe Production and Informatization (IICSPI)}, pp.\  24--27. IEEE, 2020.

\bibitem[Zou \& Topol(2025)Zou and Topol]{zou2025rise}
Zou, J. and Topol, E.~J.
\newblock The rise of agentic ai teammates in medicine.
\newblock \emph{The Lancet}, 405\penalty0 (10477):\penalty0 457, 2025.

\end{thebibliography}
\bibliographystyle{icml2024}

\newpage
\nopagebreak
\setcounter{section}{0}
\renewcommand{\thesection}{\arabic{section}} 

\section{Methods}
\subsection{Dataset and curation}\label{method:data}
In this study, we use the publicly available dataset, Semio2Brain~\cite{semio2brain}, which systematically maps seizure semiologies to brain regions through a comprehensive meta-analysis that includes 4,643 patient data constructed using PRISMA guidelines~\cite{page2021prisma} from 309 peer-reviewed publications. The dataset contains localizing data points that represent the number of patients exhibiting a reported semiology with seizure onset zones (SOZ). In particular, Semio2Brain contains 35 ictal semiological categories~\cite{beniczky2022seizureilae}, as well as a postictal and asymptomatic category, providing the most extensive public coverage of seizure manifestations to date~\cite{semio2brain}. Each semiology is linked to one or more of 103 unique brain regions, organized within seven major areas: temporal lobe, frontal lobe, cingulate gyrus, parietal lobe, occipital lobe, insula, and hypothalamus.

Each entry in the database includes a description of a seizure symptom, either a behavioral or sensory observation during a seizure, and is assigned to one or more of the seven major brain regions. The assignment of brain regions to seizure descriptions is based on two types of information:
(i) Post-operative Seizure Freedom: Knowledge about seizure freedom after resection of the brain region and (ii) simultaneously Recorded Seizure Activity: Seizure patterns recorded from intracranial or surface-based EEG located within the brain region. Both types of information serve as potential ground truths linking seizure semiology to SOZ in clinical practice. 

For our task, we focus on cases based on post-operative seizure freedom, as this is considered the gold standard for post hoc evaluation of successful SOZ identification. To effectively leverage the Semio2Brain dataset for our task, we perform several data preprocessing steps, including expanding abbreviations in the semiology descriptions, correcting spelling errors, and removing uninformative words and keywords as shown in~\ref{extfig:dataset}. This refinement process resulted in a final dataset of 1,269 reported semiology entries, each linked to one or more of the seven major brain regions for SOZ localization tasks.

\subsection{Task Formulation}
Mathematically, we task a pretrained LLM to estimate a structured likelihood distribution L across predefined brain regions given seizure semiology. Specifically, the input prompt $\widehat{P}$ comprises persona P, the user query Q, and the instruction format I for a given semiology S. And we obtain a dictionary output D such that:
\begin{equation}
    D = { r: L(r \mid S, P, Q, I) } \quad \forall r \in R
\end{equation}

where r is the key and likelihood is the value and R$=$\{``Temporal Lobe'', ``Frontal Lobe'', ``Cingulate Gyrus'', ``Parietal Lobe'', ``Occipital Lobe'', ``Insula'', ``Hypothalamus''\}. For chain-of-thought (CoT) prompting strategies, the models additionally provide structured reasoning steps and source citations formatted within the same dictionary structure.

\subsection{Prompt Strategies}
\textbf{Zero-Shot prompting:} We design a structured prompt for the aforementioned complex task, effective for zero-shot inference, where the model is expected to perform a task based solely on its pre-existing knowledge without any task-specific examples or additional training.

\textbf{Few-shot prompting:} Following the in-context learning approach described by Brown et al.~\cite{brown2020language} and Dong et al.~\cite{dong2022survey}, we incorporate K=5 examples within the input context to demonstrate the expected input-output structure. This provides the model with representative cases without requiring fine-tuning or retraining. We make use of the term few-shot (FS) and in-context learning (ICL) interchangeably in the manuscript.

\textbf{Chain-of-Thought (CoT) prompting:} Chain-of-thought prompting is a technique to ask the model to think \emph{step-by-step} and provide intermediate reasoning and sources used to get to the final answer. This technique mimics a human cognitive process, to break complex problems into small, manageable steps. It is helpful where a straightforward answer may not be trivial~\cite{wei2022chain}. To do this, we add a sentence \emph{``Solve the problem in step by step manner''} in the instruction along with the keys ``Reasoning'' and ``Sources''. 

\textbf{Few-shot CoT:} We also use a hybrid technique to combine few-shot and chain-of-thought prompting. Specifically, by providing exemplars curated by an epileptologist, we demonstrate how an expert reasons through the problem and arrives at the final decision. The model is then expected to learn in-context and mimic this reasoning process to generate its output following the format better.

\textbf{Self Consistency (SC):} Wang et al.~\cite{wang2022self} introduced the concept of ``self-consistency'', which involves generating multiple independent response chains for each query and selecting the most consistent response as the final answer. This is done via a majority voting technique. By cross-referencing different outputs, we can ensure that the final response is robust and dependable. In our problem statement, we get likelihoods of $7$ brain regions from five reasoning chains. We use median-based majority voting to get the most consistent output. For each brain region $r_i$, compute the median of its likelihoods across five iterations $L_{ij}$ for $j \in \{1, 2, 3, 4, 5\}$, then calculate adjusted likelihoods $A_{ij}$ as the absolute difference from this median:
\begin{equation}
    A_{ij} = |L_{ij} - \text{median}(L_{i1}, L_{i2}, L_{i3}, L_{i4}, L_{i5})|
\label{sub-med}
\end{equation}
We then identify the winning iteration $W$ as the one with the minimum sum of adjusted likelihoods across all seven brain regions:
\begin{equation}
    W =  \arg\min_j \sum_{i=1}^{7} A_{ij}
\label{win-med}
\end{equation}
The equations~\ref{sub-med} and~\ref{win-med} capture the entire process of majority voting using median activation. The rationale behind using the median is its robustness to outliers and its representation of the central tendency of the data. In contrast, the mean can be skewed by anomalous data, while the mode, representing the most frequent value, might fail to yield a clear result if all outputs differ.

\subsection{LLMs}
In our study, we evaluate a diverse set of large language models (LLMs), both open and proprietary models, on their effectiveness in seizure semiology localization tasks. The models are chosen to represent a range of architectures, parameter sizes, and training methodologies, allowing us to assess the state-of-the-art advancements in natural language processing for specialized medical tasks.

\textbf{Mixtral-8x7B}, developed by Mistral AI (2023)~\cite{jiang2024mixtral}, is a sparse mixture-of-experts (MoE) model. This 46.7B parameters decoder-only model employs a unique architecture: each feedforward block selects from 8 distinct expert groups, but only 12.9B parameters are used per token. A router network dynamically selects two experts per token at each layer, combining their outputs additively. As a result, Mixtral performs with the speed and cost efficiency of a 12.9B model while retaining the flexibility of a larger model. The model is pre-trained on open web data, simultaneously optimizing both experts and router networks. \textbf{Qwen-72B}  proposed by Alibaba Cloud, comprises 72 billion parameters and follows a transformer-based, decoder-only architecture. We specifically utilize Qwen-72B-Chat model\footnote{\url{https://huggingface.co/Qwen/Qwen-72B-Chat}}, renowned for its stable 32,000-token context capacity, allowing comprehensive processing of extensive textual inputs. We also incorporated \textbf{LlaMA} models developed by Meta, which have gained widespread recognition. Specifically, we test the LlaMA-70B-chat versions from both v2(released in mid 2023) and LlaMA v3 (released in 2024). Additionally, we include proprietary models from OpenAI's \textbf{GPT} series. Specifically, we use the gpt-3.5-turbo-1106 model, which represents an advanced iteration of the GPT-3.5 lineage, known for its improved responsiveness and performance over previous versions. Furthermore, we assess gpt-4-1106-preview, the latest available model from OpenAI at the time of experimentation, known for superior comprehension, nuanced reasoning, and human-like text generation capabilities. For all the models we use a temperature of 0.2. 

Recently introduced medical models fine-tuned on LlaMA-3 70B version including \textbf{OpenBioLLM-70B}\footnote{\url{https://huggingface.co/aaditya/Llama3-OpenBioLLM-70B}} (developed by Saama AI Labs) and \textbf{Med42-70B}~\cite{christophe2024med42} (developed by M42 Health AI Team) are also included in our experiments. These models showcase enhanced capabilities compared to base LlaMA models, however certain models did not follow specialized task instructions, exhibited a tendency to repeat themselves~\cite{chen2023meditron}, sensitivity to minor changes in the prompt (such as spacing and punctuation) thus lacking flexibility, or restricted context length to accommodate comprehensive symptom descriptions in the prompt~\cite{bolton2024biomedlm}. 

\subsection{Output Generation and Parsing}
To ensure standardized and structured outputs, we explicitly instruct LLMs to generate responses in a predefined JSON format, detailing likelihood percentages for seizure onset zones (SOZ) across brain regions (e.g., ``Temporal Lobe": a\%, ``Frontal Lobe": b\%). A custom regex-based parser validates format integrity and extracts key information, including SOZ likelihood estimates, reasoning chains, and literature citations. This structured approach minimizes variability in the format and improves consistency across responses. In case models returned missing, non-numeric, or ambiguous values (such as ``None"), a likelihood of 0\% is assigned for the respective brain region. 

For our multilingual evaluation, we systematically translate seizure descriptions and brain region names into French, Spanish, and Chinese using DeepL tool\footnote{\url{https://www.deepl.com}}, ensuring terminological consistency. Additionally, language-specific parsers are designed to handle these translated terms, enabling reliable structured evaluation across multiple languages while mitigating potential discrepancies arising from variations in terminology or syntax.

\subsection{Correctness Measure}
Predictions are determined by selecting the brain region with the highest likelihood value \textit{(argmax)}. These predicted regions are compared against ground truth labels to calculate precision, recall, and F1 scores for each brain region. To account for class imbalance, we use weighted averages of these metrics, with weights proportional to the number of instances in each class.

\subsection{Baselines}\label{methods:baseline}
\textbf{Naive Classifier:} As there is no established benchmark for this task, we define a lower bound performance (38.21\%) by implementing a naive classifier that always predicts the most probable class in the data. 

\textbf{Random Performance:} In addition, to assess statistically whether the models produce any meaningful outputs, we conduct a permutation-based significance test. Specifically, we create a random performance distribution by repeatedly (999 times) permuting true labels and recalculating the F1 score of each model. This procedure provides a by-chance distribution of F1 scores, capturing the expected performance if the models’ outputs were random. To quantify the deviation of our model's actual F1 score from the random distribution, we perform a Z-score normalization as follows:
\begin{equation}
    z = \frac{(x - \mu)}{\sigma}
\end{equation}

Here, x is the actual F1 score, $\mu$ is the mean F1 score obtained from the random distribution and $\sigma$ is the standard deviation of this random distribution. A high Z-score indicates significantly better performance compared to random chance. The corresponding p-values are computed to determine the statistical significance of this deviation. p-value $<$ 0.05 indicate that the actual F1 score of the LLM is unlikely to have occurred by chance, providing evidence of the model’s effectiveness beyond random performance as shown in~\ref{exttab:chance_result}.

\subsection{Confidence measure and calibration}
The likelihood output from LLMs is more informative than a single class prediction, as it allows understanding of which classes the model considers plausible, and to what degree, rather than just which class it considers the ``winner''. We leverage this feature to approximate a confidence/uncertainty measure using Shannon entropy, which is a fundamental concept in information theory that quantifies the ``fuzziness'' or uncertainty of a system's state. Given a discrete random variable $X$ with possible outcomes $x_1, x_2, \ldots, x_n$, each with probability $P(x_i)$, the Shannon entropy $H(X)$ is defined as: 
\vspace{-2mm}
\begin{equation}\label{eq:entropy}
    H(X) = -\sum\limits_{i=1}^n P(x_i)\log_2 P(x_i)
\end{equation}
The entropy ranges from 0 (100\% likelihood assigned to one brain region) to 2.807 (likelihoods uniformly distributed across all seven brain regions). We report the loss entropy, defined as (1-normalized entropy), where values approaching $1$ mean high model confidence or less uncertain and vice versa if the values tend towards $0$.

Furthermore, we evaluate model calibration~\cite{guo2017calibration} - alignment between model's predicted probabilities and empirical accuracy — using reliability diagrams and Brier scores. In this work, for each semiology $x_i$, the LLM outputs a likelihood estimate of SOZ of interest. This likelihood estimate of the true class can be represented as $p(y_i | x_i)$. To evaluate calibration across a finite set of semiologies (samples), we partition predictions into $M$ equal width bins of size $1/M$. We then compute the fraction of correct predictions within each bin, commonly known as the empirical accuracy or fraction of positives. Let $B_m$ denote the set of indices of samples whose prediction likelihood falls into the interval $I_m = (\frac{m-1}{M}, \frac{m}{M}]$ where $m \in {1,\ldots,M}$. Then the fraction of positives for bin $B_m$ is defined as:
\begin{equation}
    F_{pos}(B_m) = \frac{1}{\mid B_m \mid}\sum_{i\in B_m} 1(\hat{y_i} = y_i)
\end{equation}
where $\hat{y_i}$ and $y_i$ are the predicted and true class labels for sample i, respectively. The average confidence within a bin $B_m$ is calculated as:
\begin{equation}
    Conf(B_m) = \frac{1}{\mid B_m \mid}\sum_{i\in B_m} p(y_i | x_i)
\end{equation}
The resulting reliability diagram as shown in~\ref{fig:ext_main_res}(d) provides a visual assessment of model calibration, where perfect calibration is represented by points lying on the diagonal line $F_{pos}(B_m) = Conf(B_m)$. We quantify the calibration using Brier score~\cite{brier1950verification}, which measures the mean squared difference between predicted probabilities and actual outcomes. For N samples, the brier score $B_s$ is:
\begin{equation}
    B_s = \frac{1}{N}\sum_{j=1}^{N} (p(y_j | x_j) - y_j)^2
\end{equation}

\subsection{User Study}
We incorporated expert clinical assessment into our benchmarking framework through two structured online surveys designed to evaluate both predictive performance and reasoning quality. In the first survey, we present 81 semiologies chosen from diverse semiological categories to a clinical expert (author SL, FP). Both of these clinicians are Residents at the Department of Neurology with focus on Epileptology. The participant assigned likelihood values (0-100\%) for seizure onset zone (SOZ) localization across seven distinct brain regions using a slider interface as shown in~\ref{extfig:userstudy_form}(a). This design intentionally mirrored the output format of our computational models, enabling direct comparison between clinician and model predictions.

In the second survey [Refer~\ref{extfig:userstudy_form}(b)], we assess the quality of explanatory content generated by the best-performing models (GPT-4 and Mixtral-8x7B). To prevent bias, model identifiers were omitted from the presented reasoning. Clinical expert evaluate output for correctness and completeness across three levels. This enables us to effectively quantify the utility of long-form reasoning generated by models in epilepsy diagnosis. Additionally, two authors of the paper independently verified the accuracy of sources cited within model outputs, confirming both author lists and publication titles verbatim to ensure citation integrity.

\newpage
\nopagebreak
\setcounter{figure}{0}
\renewcommand{\thefigure}{Extended Data Figure \arabic{figure}}
\renewcommand{\figurename}{}

\setcounter{table}{0}
\renewcommand{\thetable}{Extended Data Table \arabic{table}}
\renewcommand{\tablename}{}

\begin{figure}[!ht]
\centerline{\includegraphics[width=0.8\textwidth]{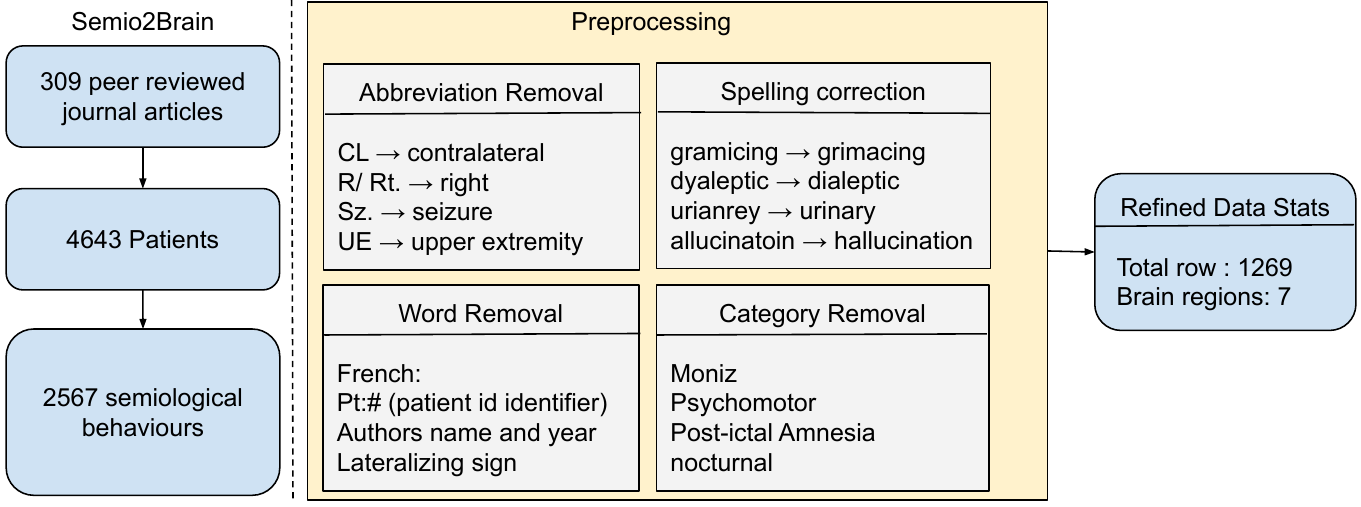}}
\caption{Data preprocessing pipeline: We use Semio2Brain~\cite{semio2brain} dataset which is a collection of 2567 semiologies spread across 7 major brain regions. Steps in preprocessing this data include removing abbreviations and replacing them with their respective full forms, correction of spelling errors present in the data, removing uninformative words and semiology categories. This result in overall 1269 rows we finally use for our analysis.}
\label{extfig:dataset}
\end{figure}

\begin{figure}[!ht]
\centerline{\includegraphics[width=0.9\textwidth]{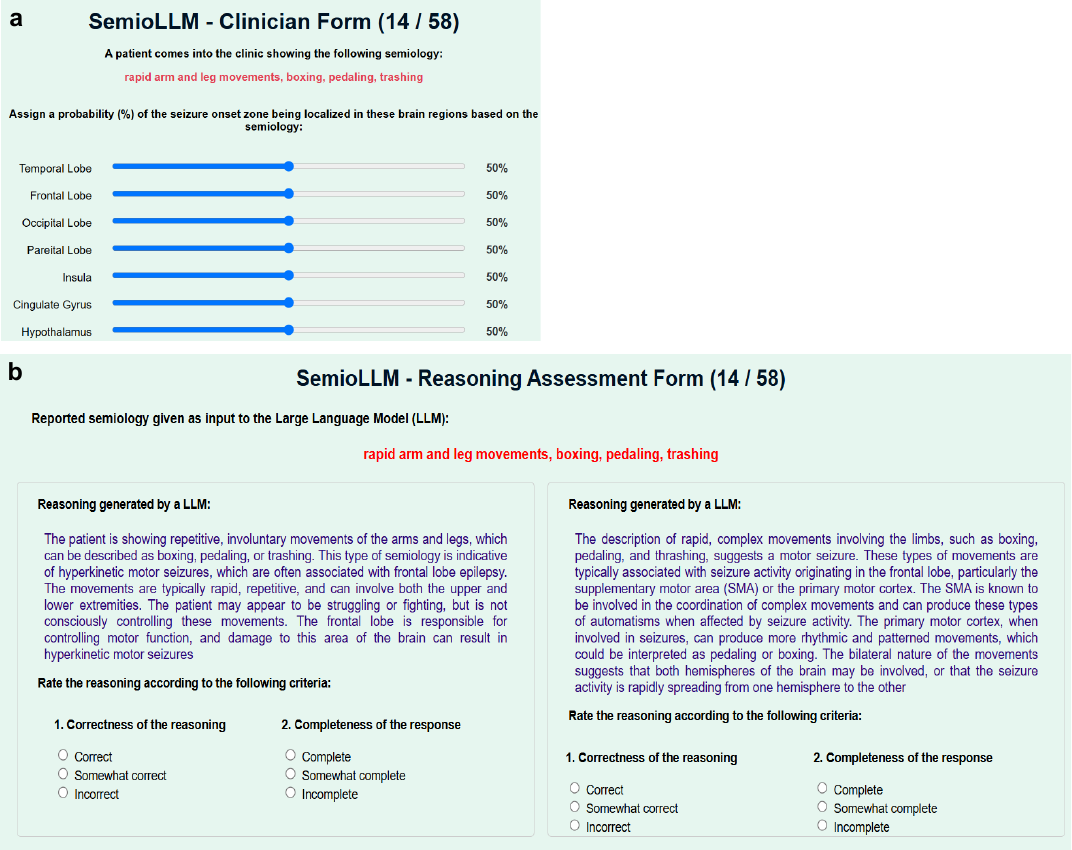}}
\caption{Clinical evaluation form illustration. (a) SOZ localization form for clinicians to provide likelihood estimates for 7 major brain regions based on 1269 semiological queries, similar to LLMs. (b) Snapshot of reasoning assessment form, where clinicians evaluate the correctness and completeness of reasoning provided by LLMs for a subset of 58 semiological queries.}
\label{extfig:userstudy_form}
\vskip -0.4in
\end{figure}

\begin{figure}[]
\begin{center}
\centerline{\includegraphics[width=0.92\textwidth]{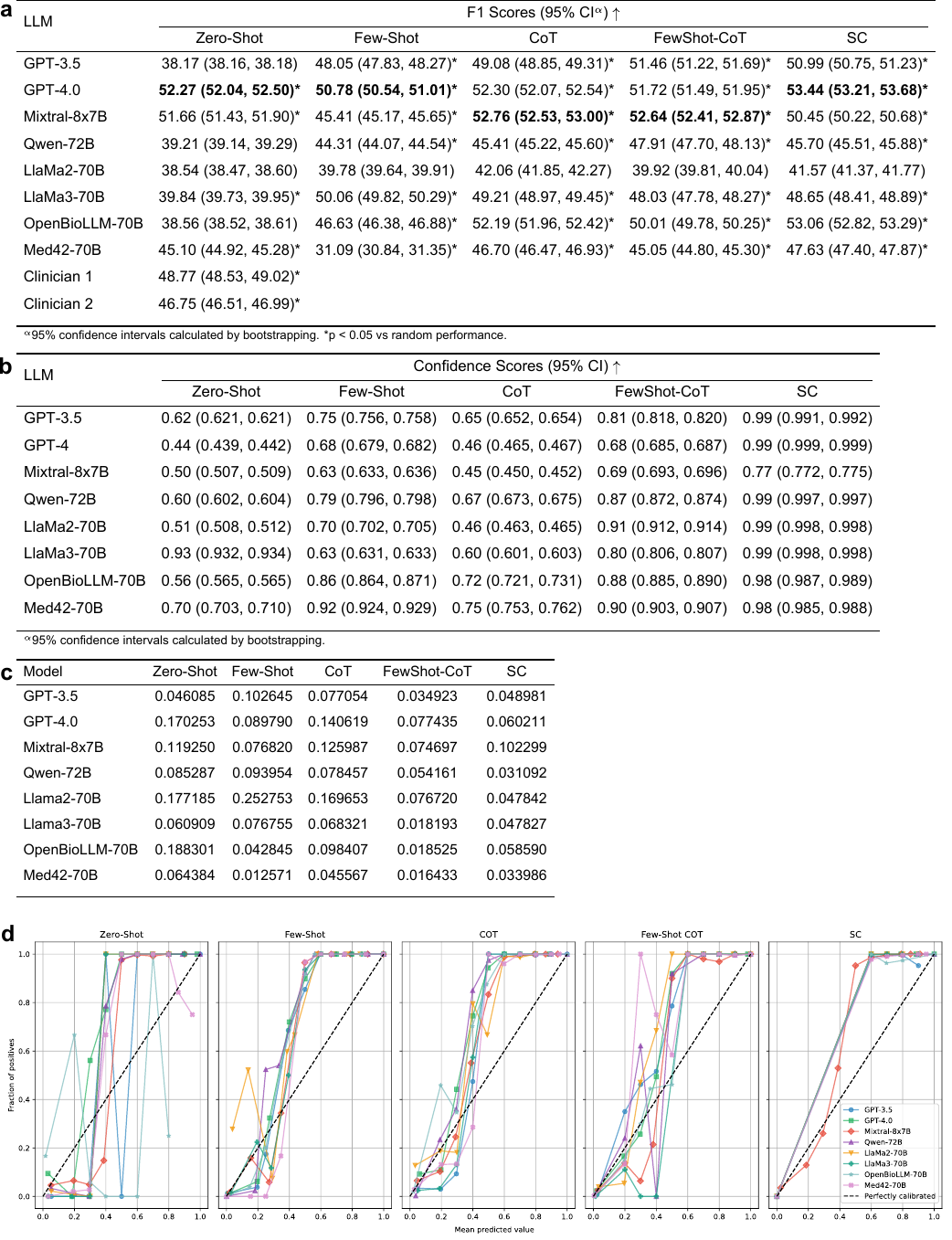}}
\caption{Data for Performance comparison of LLMs: impact of prompt engineering strategies [Zero-Shot (ZS), Few-Shot (FS), ZS-Chain-of-Thought (CoT), FS-CoT and Self Consistency (SC)]. (a) Mean F1 scores for all models with 95\% confidence intervals obtained by bootstrapping. (b) Confidence scores improve consistently with in-context learning, with FS and FS-CoT demonstrating the highest gains. (c) Brier score values to estimate calibration of each model and prompt style. (d) Reliability diagram or calibration curves with fraction of positives (actual probability) on the y-axis and predicted probability on the x-axis. The dashed black line denotes perfect calibration.}
\label{fig:ext_main_res}
\end{center}
\end{figure}

\begin{figure}[]
\centerline{\includegraphics[width=0.8\textwidth]{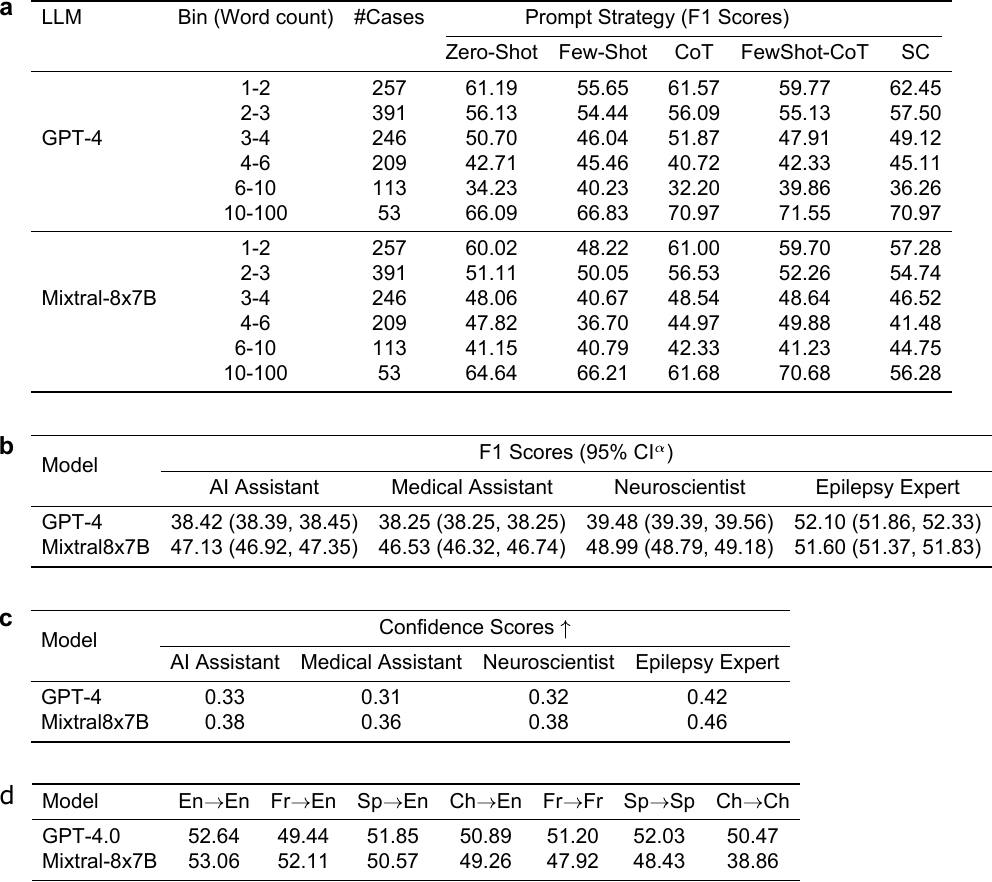}}
\caption{(a) Data table for impact of symptom description length on model performance. We bin the symptoms into 6 bins on word count and compute the F1 score across prompt styles for both GPT-4  and Mixtral-8x7B models. (b) Data Table on the influence of in-context impersonation on zero-shot performance i.e. F1 scores and (c) confidence scores computed (1 - normalized shannon entropy). Larger values represent more confidence and less uncertainty. (d) Multilingual performance analysis of GPT-4 and Mixtral-8x7B, for in-language and cross language tasks for English (En), French (Fr), Spanish (Sp) and Chinese (Ch).  (Fr→En) represents that the input semiology is in French and the output given by the model is in English and so on for others.}
\label{fig:ext_fig_ablation}
\end{figure}

\begin{figure}[]
\centerline{\includegraphics[width=\textwidth]{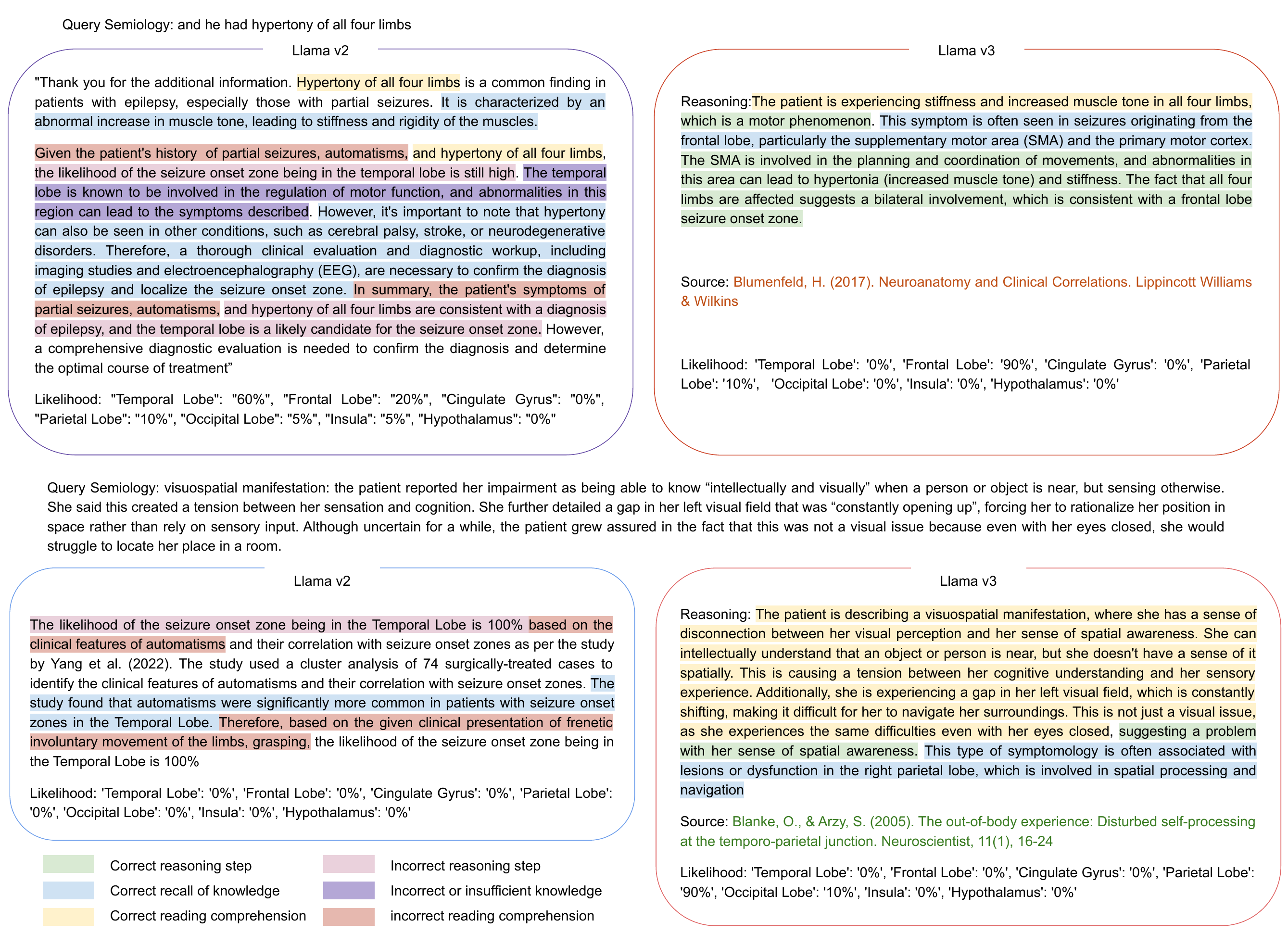}}
\caption{Generated few-shot CoT from Llama v2 70B and Llama v3 70B model on two query semiologies; including detailed annotation of comprehension, knowledge recall and reasoning capabilities with source and SOZ likelihood. We find that LlaMa v2 does not follow the instruction to give sources, show a tendency to hallucinate patient symptoms. These issues are resolved in LlaMa v3 with better instruction following capabilities and domain knowledge.}
\label{fig:reasoning_llama}
\end{figure}

\begin{table}[]
    \centering
    \begin{tabular}{llcc}
    \toprule
    \multicolumn{2}{c}{Assessment Category} & Z-value & p-value \\
        \midrule
        \multirow{3}{*}{Correctness} & Correct & 3.4893 & 0.0005* \\
        &Somewhat correct & -0.6396 & 0.5224 \\
        &Incorrect & -3.8528 & 0.0001* \\
        \\
        \multirow{3}{*}{Completeness} &Complete & 3.7715 & 0.0002* \\
        &Somewhat complete & -1.3939 & 0.1633 \\
        &Incomplete & -3.2454 & 0.0012* \\
        \bottomrule
    \end{tabular}
    \\[5pt]
        \caption{Result of two-sided z-test for proportions for correctness and completeness in Reasoning Between GPT-4 and Mixtral-8x7B models. p-values\textless0.05  indicate statistically significant differences}
    \label{tab:ztest_fig3a}
 
\end{table}

\begin{table}[]
    \centering
    
    \begin{tabular}{llcc}
        \toprule
        \multicolumn{2}{c}{Assessment Category} & Z-value & p-value \\
        \midrule
        \multirow{3}{*}{Correct} & Correct comprehension & -- & -- \\
        & Correct knowledge recall & 1.1592 & 0.2464 \\
        & Correct reasoning step & 3.8454 & 0.0001* \\
        \\
        \multirow{3}{*}{Incorrect} & Incorrect comprehension & -0.7252 & 0.4683 \\
        &Incorrect knowledge recall & -3.5920 & 0.0003* \\
        &Incorrect reasoning step & -2.4089 & 0.0160* \\
        \bottomrule
    \end{tabular}
    \\[5pt]
    \caption{Result of two-sided z-test for proportions for comparing correctness in comprehension, knowledge recall and reasoning step for FS-CoT responses from GPT-4 and Mixtral-8x7B models. p-values\textless0.05  indicate statistically significant differences}
    \label{tab:ztest_fig3b}
\end{table}

\begin{table}[]
\centering
\setlength{\tabcolsep}{5pt}
\begin{tabular}{lccccc}
\toprule
Model & Prompt Style & Mean & Std & Z-Score & P-Value \\
\midrule
\multirow{5}{*}{GPT-3.5} 
    & Zero-Shot     & 38.291459 & 0.133920 & -1.13438 & 0.283648 \\
    & Few-Shot      & 42.713399 & 1.119986 & 5.04679 & $5.89 \times 10^{-7}$ \\
    & CoT           & 42.263669 & 1.104801 & 6.38744 & $2.95 \times 10^{-10}$ \\
    & Few-Shot CoT   & 40.840135 & 1.157092 & 9.34291 & $2.62 \times 10^{-20}$ \\
    & SC            & 42.155762 & 1.099094 & 8.36114 & $1.37 \times 10^{-16}$ \\
\midrule
\multirow{5}{*}{GPT-4} 
    & Zero-Shot     & 41.567860 & 1.072080 & 10.0407 & $7.30 \times 10^{-23}$ \\
    & Few-Shot      & 39.369310 & 1.172705 & 9.88929 & $2.23 \times 10^{-22}$ \\
    & CoT           & 41.345483 & 1.142584 & 9.88593 & $2.23 \times 10^{-22}$ \\
    & Few-Shot CoT   & 39.319287 & 1.197117 & 10.6888 & $2.41 \times 10^{-25}$ \\
    & SC            & 40.700700 & 1.163469 & 11.2434 & $1.04 \times 10^{-27}$ \\
\midrule
\multirow{5}{*}{Mixtral-8x7B} 
    & Zero-Shot     & 42.701862 & 1.082395 & 8.47856 & $5.32 \times 10^{-17}$ \\
    & Few-Shot      & 38.168491 & 1.240102 & 6.00757 & $3.04 \times 10^{-9}$ \\
    & CoT           & 42.022831 & 1.141739 & 9.67015 & $1.54 \times 10^{-21}$ \\
    & Few-Shot CoT   & 41.997428 & 1.139507 & 9.62103 & $2.28 \times 10^{-21}$ \\
    & SC            & 42.235744 & 1.086164 & 7.75387 & $1.78 \times 10^{-14}$ \\
\midrule
\multirow{5}{*}{Qwen-72B} 
    & Zero-Shot     & 38.497587 & 0.295359 & 2.49959 & $1.49 \times 10^{-2}$ \\
    & Few-Shot      & 38.450994 & 1.161636 & 5.33556 & $1.37 \times 10^{-7}$ \\
    & CoT           & 40.733622 & 0.872160 & 5.43437 & $8.24 \times 10^{-8}$ \\
    & Few-Shot CoT   & 41.179159 & 1.071988 & 6.52755 & $1.22 \times 10^{-10}$ \\
    & SC            & 40.785366 & 0.850382 & 5.84517 & $7.87 \times 10^{-9}$ \\
\midrule
\multirow{5}{*}{LLaMa-v2-70B} 
    & Zero-Shot     & 38.541728 & 0.313494 & -0.22767 & 0.860897 \\
    & Few-Shot      & 39.840861 & 0.651642 & 0.00206 & 0.983596 \\
    & CoT           & 42.247646 & 1.124966 & 0.06212 & 0.973650 \\
    & Few-Shot CoT   & 39.367593 & 0.568170 & 1.17384 & 0.2729558 \\
    & SC            & 42.337105 & 1.103209 & -0.46274 & 0.6930558 \\
\midrule
\multirow{5}{*}{LLaMa-v3-70B} 
    & Zero-Shot     & 38.644281 & 0.398270 & 3.00835 & $3.24 \times 10^{-3}$ \\
    & Few-Shot      & 38.576552 & 1.170794 & 10.0374 & $7.30 \times 10^{-23}$ \\
    & CoT           & 38.542777 & 1.168171 & 9.46883 & $8.48 \times 10^{-21}$ \\
    & Few-Shot CoT   & 36.759602 & 1.163803 & 10.0417 & $7.30 \times 10^{-23}$ \\
    & SC            & 38.261547 & 1.163922 & 9.24909 & $5.59 \times 10^{-20}$ \\
\midrule
\multirow{5}{*}{OpenBioLLM-70B}
    & Zero-Shot     & 38.297745 & 0.136530 & 1.95188 & $5.94 \times 10^{-2}$ \\
    & Few-Shot      & 35.278954 & 1.180777 & 9.97077 & $1.22 \times 10^{-22}$ \\
    & CoT           & 41.947639 & 1.135520 & 9.28419 & $4.27 \times 10^{-20}$ \\
    & Few-Shot CoT   & 38.798107 & 1.188645 & 9.68262 & $1.50 \times 10^{-21}$ \\
    & SC            & 40.997045 & 1.165039 & 10.6439 & $2.60 \times 10^{-25}$ \\
\midrule
\multirow{5}{*}{Med42-70B}
    & Zero-Shot     & 41.006967 & 0.827540 & 5.17706 & $3.05 \times 10^{-7}$ \\
    & Few-Shot      & 26.119119 & 1.182827 & 4.12808 & $4.65 \times 10^{-5}$ \\
    & CoT           & 41.091527 & 1.124925 & 5.24523 & $2.18 \times 10^{-7}$ \\
    & Few-Shot CoT   & 36.853499 & 1.162500 & 7.28592 & $6.09 \times 10^{-13}$ \\
    & SC            & 41.155962 & 1.101393 & 6.06077 & $2.27 \times 10^{-9}$ \\
\midrule
Clinician 1 & - & 37.776371 & 1.192979 & 9.47082 & $8.48 \times 10^{-21}$ \\
Clinician 2
    & - & 37.422854 & 1.177575 & 8.17473 & $6.22 \times 10^{-16}$ \\
\bottomrule
\end{tabular}
\caption{Permutation test results (two-sided) showing mean, standard deviation, Z-scores, and p-values (Benjamini-Hochberg corrected) for F1 scores across different models and prompt styles.  p-value\textless0.05 shows there is a significant difference between actual model performance and random}
\label{exttab:chance_result}
\end{table}

\vspace{5mm}
\section{ML Methods}
We preprocessed each seizure semiology description using a standard NLP pipeline: tokenization (splitting text into tokens), lemmatization (reducing words to their base forms), and removal of stop words - using the Natural Language Toolkit, NLTK. This ensured that only clinically relevant content was used for downstream analysis. A dense numerical word representation is created using BERT~\cite{devlin2019bert} and ClinicalBERT~\cite{huang2019clinicalbert}, chosen to compare a general-purpose embedding and one specifically pre-trained on clinical text. Each description was embedded using both models to generate contextualized vector representations. To benchmark the effectiveness of large language models (LLMs) against traditional ML, we used the resulting embeddings as input to four classical supervised learning algorithms: Naive Bayes, Support Vector Machine (SVM), K Nearest Neighbors (KNN), and Decision Tree. Model evaluation was performed with a stratified 80/20 split on the data, repeated with 10 distinct, fixed random seeds to ensure reproducibility. The F1 score was reported as the primary metric, averaged across 10-folds. For direct comparison, we also evaluated GPT-4.0 on the same 10-fold test splits.

\begin{figure}[t]
\begin{center}
\centerline{\includegraphics[width=1.0\textwidth]{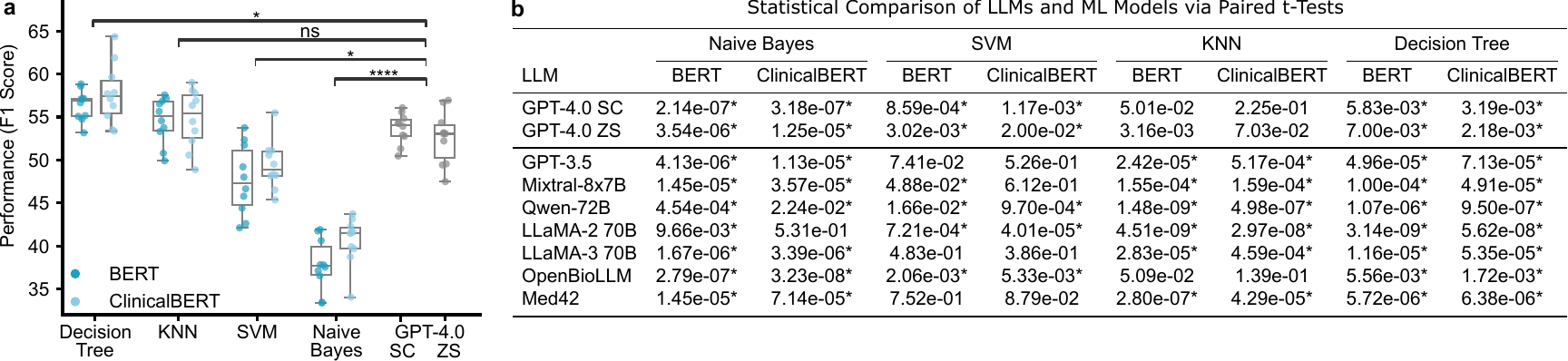}}
\caption{Benchmarking Machine Learning (ML) Algorithms for SOZ localization task: (a) F1 scores for classical ML models (Naive Bayes, Support Vector Machine (SVM), K Nearest Neighbors (KNN), Decision Tree) using BERT and ClinicalBERT embeddings as features in comparison to performance of GPT-4.0 (prompt strategy: self-consistency (SC) and zero-shot (ZS)) as an end-to-end predictor on the same task and test set. Decision Tree performed comparably to the top LLMs (GPT-4.0), KNN showed no significant difference in some cases, while Naive Bayes and SVM were significantly worse than LLMs (b) paired t-test result comparing all ML models and LLMs evaluated in this study. * denotes p$<$0.05 i.e. significant difference.}
\label{extfig:mlbaseline}
\end{center}
\end{figure}

As shown in Fig~\ref{extfig:mlbaseline}(a),ClinicalBERT, trained on medical corpora, consistently outperformed BERT trained on general text. Among classifiers, Decision Tree performed comparably to the top LLMs (GPT-4.0; mean F1 score for self-consistency: 53.52), KNN achieved a similar mean F1 score (54.61) with no statistically significant difference, while Naive Bayes (37.79) and SVM (47.63) were significantly worse. This is consistent with Fig.~\ref{extfig:mlbaseline}(b), where GPT-4.0 matched the best classical ML models, demonstrating that LLMs, when used as direct predictors, can perform on par with the strongest supervised ML pipelines leveraging handcrafted embeddings for seizure semiology classification.

However, LLMs provide important advantages over classical ML methods: they generate interpretable natural language reasoning and supporting evidence for their outputs - enhancing clinical trust, and can be rapidly adapted to new tasks or prompt structures without retraining or manual feature engineering, enabling scalable and flexible deployment for evolving clinical needs.

\end{document}